\pdfoutput=1
\pdftrailerid{}
\documentclass[11pt]{article}

\usepackage[utf8]{inputenc}
\usepackage[T1]{fontenc}
\usepackage[margin=1in]{geometry}
\usepackage[hidelinks,pdfauthor={},pdftitle={},pdfsubject={},pdfkeywords={}]{hyperref}
\usepackage{url}
\usepackage{booktabs}
\usepackage{amsfonts}
\usepackage{nicefrac}
\usepackage{microtype}
\usepackage{amsmath, amssymb, amsthm}
\usepackage{graphicx}
\usepackage{xcolor}
\usepackage{natbib}
\newcommand{\us}{\_\allowbreak}
\newcommand{\slsh}{/\allowbreak}
\graphicspath{{figs/}}

\newtheorem{theorem}{Theorem}[section]

\newtheorem{proposition}[theorem]{Proposition}
\newtheorem{observation}[theorem]{Observation}

\theoremstyle{definition}

\newcommand{\E}{\mathbb{E}}
\newcommand{\R}{\mathbb{R}}
\newcommand{\Sig}{\boldsymbol{\Sigma}}
\newcommand{\relDM}{\mathrm{rel}\text{-}d_M}
\newcommand{\sstar}{s^{\ast}}

\title{A Self-Generated Floor on Adam's $\varepsilon$: \\
  Evidence from Power-Law Matrix Sensing}

\author{Lngel Kyo\\
  A2O Labs\\
  \texttt{lngel.kyo@a2o-labs.com}}
\date{}

\newcommand{\numAuErr}{2.3}%
\newcommand{\numAuErrBare}{10.5}%
\newcommand{\numAuTrack}{1.08}%
\newcommand{\numAdamResidHi}{0.22}%
\newcommand{\numAdamResidLo}{0.05}%
\newcommand{\numAliasHi}{69}%
\newcommand{\numAliasLo}{21}%
\newcommand{\numAlignDecaySpreadFive}{10.1}%
\newcommand{\numAlignDecaySpreadSix}{1.5}%
\newcommand{\numAlignDexFive}{+0.286}%
\newcommand{\numAlignDexSix}{-0.343}%
\newcommand{\numAlignFloorExp}{0.93}%
\newcommand{\numAlignFloorLocalSlopeLast}{0.06}%
\newcommand{\numAlignFloorLocalSlopes}{1.83/0.91/0.06}%
\newcommand{\numAlignFloorLooFolds}{0.46/0.99/0.87/1.40}%
\newcommand{\numAlignFloorLooHi}{1.40}%
\newcommand{\numAlignFloorLooLo}{0.46}%
\newcommand{\numAlignGapFive}{19.6}%
\newcommand{\numAlignGapSix}{22.7}%
\newcommand{\numAlignLooFiveHarmful}{2}%
\newcommand{\numAlignLooFiveHi}{0.85}%
\newcommand{\numAlignLooFiveLo}{0.28}%
\newcommand{\numAlignLooFiveSchedule}{0}%
\newcommand{\numAlignLooSixHarmful}{0}%
\newcommand{\numAlignLooSixHi}{2.63}%
\newcommand{\numAlignLooSixLo}{1.38}%
\newcommand{\numAlignLooSixSchedule}{3}%
\newcommand{\numAlignMeanFive}{0.498}%
\newcommand{\numAlignMeanSix}{2.293}%
\newcommand{\numAlignNullSpreadFive}{22}%
\newcommand{\numAlignNullSpreadSix}{15}%
\newcommand{\numAlignPairedFive}{0.38}%
\newcommand{\numAlignPairedFiveDir}{2}%
\newcommand{\numAlignPairedFiveHi}{2.28}%
\newcommand{\numAlignPairedFiveLo}{0.059}%
\newcommand{\numAlignPairedSix}{1.59}%
\newcommand{\numAlignPairedSixDir}{4}%
\newcommand{\numAlignPairedSixHi}{4.38}%
\newcommand{\numAlignPairedSixLo}{0.35}%
\newcommand{\numAlignRatioFive}{0.551}%
\newcommand{\numAlignRatioSix}{1.365}%
\newcommand{\numAlignWorseFive}{1.81}%
\newcommand{\numAttrPeriodHi}{2.43}%
\newcommand{\numAttrPeriodLo}{2.32}%
\newcommand{\numAttrRmsHi}{1.14}%
\newcommand{\numAttrRmsLo}{1.07}%
\newcommand{\numAuConjAlphaSpan}{100}%
\newcommand{\numAuConjEpsSpan}{10000}%
\newcommand{\numAuConjSpread}{0.66}%
\newcommand{\numBaseFinalFive}{14.6}%
\newcommand{\numBaseFinalSix}{29.2}%
\newcommand{\numBetaOneArms}{5}%
\newcommand{\numBetaOneCompHi}{0.97}%
\newcommand{\numBetaOneCompLo}{0.70}%
\newcommand{\numBetaOneCompNineNine}{0.86}%
\newcommand{\numBetaOneCompZero}{0.45}%
\newcommand{\numBetaOneDirArms}{3}%
\newcommand{\numBetaOneLeverage}{55}%
\newcommand{\numBetaOnePairSpreadMax}{0.34}%
\newcommand{\numBetaOnePairSpreadTyp}{0.14}%
\newcommand{\numBetaOneRatioEight}{1.96}%
\newcommand{\numBetaOneRatioEightPred}{2.11}%
\newcommand{\numBetaOneRatioHalf}{4.42}%
\newcommand{\numBetaOneRatioHalfPred}{6.33}%
\newcommand{\numBetaOneRatioNineFive}{0.47}%
\newcommand{\numBetaOneRatioNineFivePred}{0.49}%
\newcommand{\numBetaOneRatioNineNine}{0.082}%
\newcommand{\numBetaOneRatioNineNinePred}{0.095}%
\newcommand{\numBetaOneSlope}{0.97}%
\newcommand{\numBetaOneSlopeDropOne}{0.84}%
\newcommand{\numBetaOneSpan}{66}%
\newcommand{\numBetaOneWallEight}{5.04}%
\newcommand{\numBetaOneWallHalf}{4.19}%
\newcommand{\numBetaOneWallNine}{5.42}%
\newcommand{\numBetaOneWallNineFive}{5.84}%
\newcommand{\numBetaOneWallNineNine}{6.60}%
\newcommand{\numBetaOneWallZero}{2.84}%
\newcommand{\numBootCensoredSixHundredK}{547}%
\newcommand{\numBootCensoredSixtyK}{0}%
\newcommand{\numBootCensoredTwoHundredK}{0}%
\newcommand{\numBootDisjointSixHundredK}{6.16/6.07/5.21/5.26}%
\newcommand{\numBootDisjointSixtyK}{5.42/5.81/5.42/5.99}%
\newcommand{\numBootDisjointSpreadSixHundredK}{0.955}%
\newcommand{\numBootDisjointSpreadSixtyK}{0.569}%
\newcommand{\numBootDisjointSpreadTwoHundredK}{0.516}%
\newcommand{\numBootDisjointTwoHundredK}{5.65/5.76/5.59/6.11}%
\newcommand{\numBootHiSixHundredK}{6.30}%
\newcommand{\numBootHiSixtyK}{5.99}%
\newcommand{\numBootHiTwoHundredK}{6.14}%
\newcommand{\numBootLoSixHundredK}{5.07}%
\newcommand{\numBootLoSixtyK}{5.37}%
\newcommand{\numBootLoTwoHundredK}{5.55}%
\newcommand{\numBootNSixHundredK}{15504}%
\newcommand{\numBootNSixtyK}{15504}%
\newcommand{\numBootNTwoHundredK}{15504}%
\newcommand{\numBootOverLoo}{1.6}%
\newcommand{\numBootRangeOverLoo}{3.6}%
\newcommand{\numBootRangeSixHundredK}{1.235}%
\newcommand{\numBootRangeSixtyK}{0.625}%
\newcommand{\numBootRangeTwoHundredK}{0.590}%
\newcommand{\numBootSdSixHundredK}{0.43}%
\newcommand{\numBootSdSixtyK}{0.14}%
\newcommand{\numBootSdTwoHundredK}{0.18}%
\newcommand{\numBoundaryHi}{0.012}%
\newcommand{\numBoundaryLo}{0.005}%
\newcommand{\numBoundaryU}{38.3}%
\newcommand{\numCensorVOneSixDEightSixGridMax}{3.0}%
\newcommand{\numCreepRatio}{1.22}%
\newcommand{\numCreepRatioRare}{2.09}%
\newcommand{\numCycleRows}{227}%
\newcommand{\numDecadeGapSixHundredK}{0.884}%
\newcommand{\numDecadeGapSixtyK}{0.029}%
\newcommand{\numDecadeGapTwoHundredK}{0.197}%
\newcommand{\numDecadeOneSixHundredK}{6.125}%
\newcommand{\numDecadeOneSixtyK}{5.613}%
\newcommand{\numDecadeOneTwoHundredK}{5.689}%
\newcommand{\numDecadeTwoSixHundredK}{5.240}%
\newcommand{\numDecadeTwoSixtyK}{5.584}%
\newcommand{\numDecadeTwoTwoHundredK}{5.887}%
\newcommand{\numDecayFinalFive}{0.5}%
\newcommand{\numDecayFinalSix}{8.3}%
\newcommand{\numDecayGainFive}{31.2}%
\newcommand{\numDecayGainSix}{3.5}%
\newcommand{\numDecayPairedSixA}{6.5}%
\newcommand{\numDecayPairedSixB}{57.5}%
\newcommand{\numDecayPairedSixC}{1.5}%
\newcommand{\numDecaySepHi}{224}%
\newcommand{\numDecaySepLo}{53}%
\newcommand{\numDefaultLrAbove}{1}%
\newcommand{\numDefaultLrLooHi}{0.46}%
\newcommand{\numDefaultLrLooLo}{0.21}%
\newcommand{\numDefaultLrPairRatio}{8.9}%
\newcommand{\numDefaultLrRhoHi}{1.26}%
\newcommand{\numDefaultLrRhoLo}{0.09}%
\newcommand{\numDefaultLrRhoMedian}{0.22}%
\newcommand{\numDefaultLrSpread}{14}%
\newcommand{\numDeltaBranch}{1.35}%
\newcommand{\numDeltaFree}{0.67}%
\newcommand{\numDutyMargin}{0.0067}%
\newcommand{\numDutyPostConst}{0.172}%
\newcommand{\numDutyPostDecay}{0.000}%
\newcommand{\numDutyPostDecayMargin}{3.3}%
\newcommand{\numDutyPostDecayPeak}{0.04836}%
\newcommand{\numDutySFourZero}{0.00}%
\newcommand{\numDutySFourFive}{0.00}%
\newcommand{\numDutySFiveZero}{0.02}%
\newcommand{\numDutySFiveFive}{0.13}%
\newcommand{\numDutySSixZero}{0.39}%
\newcommand{\numDutySpan}{0.2433}%
\newcommand{\numDutySpanApprox}{0.24}%
\newcommand{\numDutySpanLooCross}{2}%
\newcommand{\numDutySpanLooHi}{0.308}%
\newcommand{\numDutySpanLooLo}{0.204}%
\newcommand{\numDutySpanLooN}{5}%
\newcommand{\numDutyWall}{5.632}%
\newcommand{\numDutyWallHi}{5.493}%
\newcommand{\numDutyWallLo}{5.736}%
\newcommand{\numEpsBranchFactor}{21}%
\newcommand{\numEpsBranchOffset}{+0.87}%
\newcommand{\numEpsBranchSlope}{1.21}%
\newcommand{\numEpsDyn}{4.76}%
\newcommand{\numEpsDynAll}{1.87}%
\newcommand{\numEpsDynBorrowedOverKappaAlpha}{0.08}%
\newcommand{\numEpsDynBranch}{210}%
\newcommand{\numEpsDynBranchOverKappaAlpha}{3.7}%
\newcommand{\numEpsDynDeltaOne}{45}%
\newcommand{\numEpsDynFree}{10.8}%
\newcommand{\numEpsDynHi}{10.29}%
\newcommand{\numEpsDynLo}{4.76}%
\newcommand{\numEpsDynLog}{3.93}%
\newcommand{\numEpsDynLogPct}{-17}%
\newcommand{\numEpsDynRatio}{2.16}%
\newcommand{\numEpsDynRerun}{4.68}%
\newcommand{\numEpsDynRerunOverStored}{0.98}%
\newcommand{\numEpsSweepCount}{6}%
\newcommand{\numEpsSweepHi}{10^{-2}}%
\newcommand{\numEpsSweepLo}{10^{-12}}%
\newcommand{\numExtMedHi}{1.25}%
\newcommand{\numExtMedLo}{0.91}%
\newcommand{\numExtWithin}{71.8}%
\newcommand{\numFRerunDiffer}{102}%
\newcommand{\numFRerunExact}{18}%
\newcommand{\numFRerunFiles}{29}%
\newcommand{\numFRerunOnset}{4}%
\newcommand{\numFRerunReads}{32}%
\newcommand{\numFRerunSameBranch}{7}%
\newcommand{\numFamTwoDelta}{+0.22}%
\newcommand{\numFamTwoDeltaLooHi}{+0.19}%
\newcommand{\numFamTwoDeltaLooLo}{+0.02}%
\newcommand{\numFamTwoKappa}{5.4}%
\newcommand{\numFamTwoOnsetTen}{7.68}%
\newcommand{\numFamTwoOnsetTwelve}{7.90}%
\newcommand{\numFamTwoRhoHi}{100}%
\newcommand{\numFamTwoRhoLo}{26}%
\newcommand{\numFamTwoRhoMedian}{76}%
\newcommand{\numFigAboveCells}{4}%
\newcommand{\numFigAgreePct}{6}%
\newcommand{\numFloorConst}{3.5}%
\newcommand{\numFloorDecades}{1.5}%
\newcommand{\numFloorDecayOneM}{5.5}%
\newcommand{\numFloorExp}{1.85}%
\newcommand{\numFloorExpDecayed}{1.95}%
\newcommand{\numFloorOneE}{5.9}%
\newcommand{\numFloorRatio}{63}%
\newcommand{\numFloorRatioFiveHundred}{59}%
\newcommand{\numFloorThreeE}{4.9}%
\newcommand{\numFloorThreeF}{5.4}%
\newcommand{\numFloqCrossHi}{0.1724}%
\newcommand{\numFloqCrossLo}{0.1716}%
\newcommand{\numFloqGridN}{42}%
\newcommand{\numFloqRepelCount}{27}%
\newcommand{\numFloqZeroMuMid}{0.999}%
\newcommand{\numGapOverInstabilityHi}{1.1}%
\newcommand{\numGapOverInstabilityLo}{0.8}%
\newcommand{\numInterpBracketRatioVThreeTwoDFour}{7.0}%
\newcommand{\numInterpCellsExceedLoo}{1}%
\newcommand{\numInterpShiftHi}{+0.087}%
\newcommand{\numInterpShiftLo}{+0.021}%
\newcommand{\numInterpShiftMean}{+0.044}%
\newcommand{\numKapCellVOneSixDTwo}{0.018}%
\newcommand{\numKapCellVOneSixDFour}{0.51}%
\newcommand{\numKapCellVOneSixDEight}{0.53}%
\newcommand{\numKapCellVThreeTwoDFour}{0.10}%
\newcommand{\numKapCellVThreeTwoDEight}{0.047}%
\newcommand{\numKapCellVEightDFour}{2.77}%
\newcommand{\numKapCfgPlateauShift}{0.41}%
\newcommand{\numKapCfgRatio}{0.14}%
\newcommand{\numKapCfgSixteenEight}{0.38}%
\newcommand{\numKapDeltaMeasVOneSixDTwo}{+1.10}%
\newcommand{\numKapDeltaMeasVOneSixDFour}{-0.03}%
\newcommand{\numKapDeltaMeasVOneSixDEight}{+0.06}%
\newcommand{\numKapDeltaMeasVThreeTwoDFour}{-0.15}%
\newcommand{\numKapDeltaMeasVThreeTwoDEight}{+0.15}%
\newcommand{\numKapDeltaMeasVEightDFour}{+0.00}%
\newcommand{\numKapDeltaPredVOneSixDTwo}{+1.45}%
\newcommand{\numKapDeltaPredVOneSixDFour}{+0.41}%
\newcommand{\numKapDeltaPredVOneSixDEight}{+0.34}%
\newcommand{\numKapDeltaPredVThreeTwoDFour}{+0.68}%
\newcommand{\numKapDeltaPredVThreeTwoDEight}{+0.74}%
\newcommand{\numKapDeltaPredVEightDFour}{+0.00}%
\newcommand{\numKapDiscCells}{4}%
\newcommand{\numKapLiteralPassCells}{1}%
\newcommand{\numKapRatioVOneSixDTwo}{0.0066}%
\newcommand{\numKapRatioVOneSixDFour}{0.18}%
\newcommand{\numKapRatioVOneSixDEight}{0.19}%
\newcommand{\numKapRatioVThreeTwoDFour}{0.037}%
\newcommand{\numKapRatioVThreeTwoDEight}{0.017}%
\newcommand{\numKapRatioVEightDFour}{1.00}%
\newcommand{\numKapTrackVerdict}{WALL-EQUIVALENT-SCALAR}%
\newcommand{\numKapVotesScalar}{4}%
\newcommand{\numKapVotesTracks}{0}%
\newcommand{\numKappa}{5.65}%
\newcommand{\numKappaCurrentSix}{2.77}%
\newcommand{\numKappaGap}{12}%
\newcommand{\numKappaSCells}{7}%
\newcommand{\numKappaSSpread}{27}%
\newcommand{\numKappaStoredOverCurrent}{2.0}%
\newcommand{\numLadderFracEightTen}{5.7}%
\newcommand{\numLadderFracFourSix}{91}%
\newcommand{\numLadderFracSixEight}{49}%
\newcommand{\numLadderFracTenTwelve}{-0.6}%
\newcommand{\numLadderShiftMax}{+0.112}%
\newcommand{\numLbfgsConverged}{11}%
\newcommand{\numLbfgsN}{12}%
\newcommand{\numLbfgsRelMax}{5.3}%
\newcommand{\numLbfgsStepsHi}{96}%
\newcommand{\numLbfgsStepsLo}{7.9}%
\newcommand{\numLooHalfWidthVEightDFour}{0.086}%
\newcommand{\numLrArmShiftMax}{0.016}%
\newcommand{\numMbDeltaTwelve}{+0.06}%
\newcommand{\numMbDeltaTwentyFour}{-0.30}%
\newcommand{\numMbLooHiTwelve}{+0.27}%
\newcommand{\numMbLooHiTwentyFour}{-0.22}%
\newcommand{\numMbLooLoTwelve}{-0.06}%
\newcommand{\numMbLooLoTwentyFour}{-0.43}%
\newcommand{\numMbRhoBothTwelve}{6.91}%
\newcommand{\numMbRhoBothTwentyFour}{3.01}%
\newcommand{\numMbRhoTwelve}{6.9}%
\newcommand{\numMbRhoTwentyFour}{3.0}%
\newcommand{\numMbWallTwelve}{3.84}%
\newcommand{\numMbWallTwentyFour}{4.44}%
\newcommand{\numMuHi}{1.02}%
\newcommand{\numMuLo}{1.31}%
\newcommand{\numMuMid}{1.10}%
\newcommand{\numNullGainFive}{4.3}%
\newcommand{\numNullGainFiveOneE}{3.2}%
\newcommand{\numNullGainFiveThreeE}{5.4}%
\newcommand{\numNullGainSix}{3.2}%
\newcommand{\numNullGainSixOneE}{3.2}%
\newcommand{\numNullGainSixThreeE}{3.2}%
\newcommand{\numOnsetEpsEight}{5.339}%
\newcommand{\numOnsetEpsFour}{3.47}%
\newcommand{\numOnsetEpsSix}{4.68}%
\newcommand{\numOnsetEpsTen}{5.414}%
\newcommand{\numOnsetEpsTwelve}{5.407}%
\newcommand{\numOnsetEpsTwo}{2.01}%
\newcommand{\numOnsetTwentySixHundredK}{5.56}%
\newcommand{\numOnsetTwentySixtyK}{5.60}%
\newcommand{\numOnsetTwentyTwoHundredK}{5.76}%
\newcommand{\numOnsetVOneSixDTwo}{6.315}%
\newcommand{\numOnsetVOneSixDFour}{4.391}%
\newcommand{\numOnsetVOneSixDEight}{3.923}%
\newcommand{\numOnsetVThreeTwoDFour}{3.571}%
\newcommand{\numOnsetVThreeTwoDEight}{3.459}%
\newcommand{\numOnsetVEightDFour}{5.407}%
\newcommand{\numOnsetVOneSixDEightFour}{2.242}%
\newcommand{\numOnsetVOneSixDEightSix}{3.228}%
\newcommand{\numOnsetVOneSixDEightTen}{3.947}%
\newcommand{\numOnsetVThreeTwoDFourLog}{3.658}%
\newcommand{\numOnsetVThreeTwoDFourLooWidth}{0.036}%
\newcommand{\numPNineLooAboveLog}{2}%
\newcommand{\numPNineOnsetLog}{5.986}%
\newcommand{\numPOneDelta}{0.045}%
\newcommand{\numPathDeltaSixtyK}{+0.004}%
\newcommand{\numPathDeltaSixtyKLooHi}{+0.0045}%
\newcommand{\numPathDeltaSixtyKLooLo}{-0.0001}%
\newcommand{\numPathOnsetTen}{5.419}%
\newcommand{\numPathOnsetTwelve}{5.423}%
\newcommand{\numPathTenDeltaSixHundredK}{+0.009}%
\newcommand{\numPathTenDeltaSixHundredKLooHi}{+0.11}%
\newcommand{\numPathTenDeltaSixHundredKLooLo}{-0.12}%
\newcommand{\numPathTenDeltaSixtyK}{-0.007}%
\newcommand{\numPathTenDeltaSixtyKLooHi}{+0.00}%
\newcommand{\numPathTenDeltaSixtyKLooLo}{-0.02}%
\newcommand{\numPathTenDeltaTwoHundredK}{-0.129}%
\newcommand{\numPathTenDeltaTwoHundredKLooHi}{+0.02}%
\newcommand{\numPathTenDeltaTwoHundredKLooLo}{-0.28}%
\newcommand{\numPathTenOnsetTenSixHundredK}{6.12}%
\newcommand{\numPathTenOnsetTenSixtyK}{5.62}%
\newcommand{\numPathTenOnsetTenTwoHundredK}{5.82}%
\newcommand{\numPathTenOnsetTwelveSixHundredK}{6.12}%
\newcommand{\numPathTenOnsetTwelveSixtyK}{5.61}%
\newcommand{\numPathTenOnsetTwelveTwoHundredK}{5.69}%
\newcommand{\numPlateauResidVOneSixDEight}{0.06}%
\newcommand{\numPredEpsTwoExtrap}{2.14}%
\newcommand{\numPredVOneSixDTwo}{5.21}%
\newcommand{\numPredVOneSixDFour}{4.42}%
\newcommand{\numPredVOneSixDEight}{3.86}%
\newcommand{\numPredVThreeTwoDFour}{3.72}%
\newcommand{\numPredVThreeTwoDEight}{3.31}%
\newcommand{\numPredVEightDFour}{5.41}%
\newcommand{\numPredVOneSixDEightSixExtrap}{3.191}%
\newcommand{\numRerunCells}{3}%
\newcommand{\numRerunMaxShift}{0.22}%
\newcommand{\numRerunOutsideLoo}{2}%
\newcommand{\numRerunReproduced}{3}%
\newcommand{\numRerunShiftVOneSixDFour}{+0.124}%
\newcommand{\numRerunShiftVOneSixDFourLog}{+0.097}%
\newcommand{\numRerunShiftVThreeTwoDFour}{+0.079}%
\newcommand{\numRerunShiftVThreeTwoDFourLog}{+0.072}%
\newcommand{\numRerunShiftVEightDFour}{+0.016}%
\newcommand{\numRerunShiftVEightDFourLog}{+0.010}%
\newcommand{\numRerunShiftVOneSixDEight}{-0.034}%
\newcommand{\numRerunShiftVOneSixDTwo}{+0.001}%
\newcommand{\numRerunShiftVThreeTwoDEight}{-0.220}%
\newcommand{\numRerunVOneSixDFour}{4.515}%
\newcommand{\numRerunVThreeTwoDFour}{3.650}%
\newcommand{\numRerunVEightDFour}{5.423}%
\newcommand{\numRerunVOneSixDEight}{3.889}%
\newcommand{\numRerunVOneSixDTwo}{6.316}%
\newcommand{\numRerunVThreeTwoDEight}{3.239}%
\newcommand{\numRerunVThreeTwoDFourLog}{3.730}%
\newcommand{\numResidDtwo}{1.10}%
\newcommand{\numRmsAll}{0.402}%
\newcommand{\numRmsAmp}{1.10}%
\newcommand{\numRmsDge}{0.098}%
\newcommand{\numRmsDgeBranch}{0.146}%
\newcommand{\numRmsDgeDeltaOne}{0.107}%
\newcommand{\numRmsDgeFree}{0.093}%
\newcommand{\numRmsDgeLog}{0.081}%
\newcommand{\numRmsDgeRerun}{0.062}%
\newcommand{\numRmsSpanHi}{0.098}%
\newcommand{\numRmsSpanLo}{0.093}%
\newcommand{\numRmsVDFree}{0.137}%
\newcommand{\numRmspropPairs}{15}%
\newcommand{\numRmspropPred}{19}%
\newcommand{\numRmspropRatio}{8.5}%
\newcommand{\numRowsInBand}{45.4}%
\newcommand{\numSatMoveMeasured}{-0.007}%
\newcommand{\numSatMoveMeasuredAbs}{0.007}%
\newcommand{\numSatMoveMeasuredB}{-0.024}%
\newcommand{\numSatMoveMeasuredLog}{+0.000}%
\newcommand{\numSatMovePredicted}{1.33}%
\newcommand{\numSatMovePredictedB}{0.95}%
\newcommand{\numSelfEps}{5.65}%
\newcommand{\numSoftmaxSlopeErrPct}{3.6}%
\newcommand{\numSoftmaxSlopeMeasured}{0.4919}%
\newcommand{\numSoftmaxSlopePredicted}{0.4746}%
\newcommand{\numWallKappaAlphaBorrowed}{4.70}%
\newcommand{\numWallKappaAlphaBranchResidVEightDFour}{+0.16}%
\newcommand{\numWallKappaAlphaBranchResidVOneSixDEight}{+0.30}%
\newcommand{\numWallKappaAlphaBranchVEightDFour}{5.57}%
\newcommand{\numWallKappaAlphaBranchVOneSixDEight}{4.22}%
\newcommand{\numWorstInstability}{1.235}%

\begin{document}
\maketitle

\begin{abstract}
In nonconvex Burer--Monteiro low-rank matrix sensing of $U^\ast \in \R^{V \times D}$ under power-law covariance $\Sig = \mathrm{diag}(j^{-s})_{j=1}^V$, the precision-wall approximation predicts that Adam stalls when the smallest non-gauge Hessian eigenvalue drops below its denominator floor, giving $\sstar(V,D) = [\log(1/\varepsilon_{\mathrm{eff}}) - \log V]/\log(VD) + \delta_{\mathrm{Adam}}$. If $\varepsilon_{\mathrm{eff}}$ were set by $\varepsilon_0$, lowering $\varepsilon_0$ from $10^{-10}$ to $10^{-12}$ would raise $\sstar$ by $\numSatMovePredictedB\text{--}\numSatMovePredicted$. Instead, across six values of $\varepsilon_0$ two decades apart, $10^{-2}$ to $10^{-12}$, the boundary follows the $\varepsilon_0$ slope down to $10^{-6}$, bends near $10^{-8}$, and is flat below $10^{-8}$, while warm-started L-BFGS and Gauss--Newton reach machine precision on stalled instances. We trace this saturation to a self-generated dynamical floor $\varepsilon_{\mathrm{dyn}} \propto \alpha$: high-curvature modes oscillate persistently near the adaptive edge of stability with amplitude $\propto \alpha(1-\beta_1)/(1+\beta_1)$ and leak gradient fluctuations of scale $\kappa\alpha$ into rare modes via measurement coupling, pinning $\sqrt{\hat{v}_V} \approx \kappa\alpha \gg \varepsilon_0$. Fitting a single threshold-crossing wall parameter $\varepsilon_{\mathrm{dyn}} = \numEpsDyn\times 10^{-9}$ (with the detection lag frozen at its companion calibration) predicts the saturated failure boundary for $D \ge 4$ across five $(V,D)$ scales to RMS $\numRmsDge$ (breaking down at $D = 2$), while shrinking $\varepsilon_0$ below $10^{-8}$ is ineffective and lowering the step size $\alpha$ reduces the floor directly.
\end{abstract}

\section{Introduction}
\label{sec:intro}

When optimizing nonconvex models with heavy-tailed feature spectra, practitioners routinely treat Adam's denominator constant $\varepsilon_0$ (default $10^{-8}$) as a passive guard against division by zero, or tune it downward when small gradients underflow~\citep{kingma2015adam,everett2024adam}. In our companion study of rank-$D$ Burer--Monteiro matrix sensing ($M^\ast = U^\ast {U^\ast}^\top \in \R^{V \times V}$, $U^\ast \in \R^{V \times D}$) from $m = 1.5VD$ linear measurements $A_k = \Sig^{1/2} G_{\mathrm{sym},k} \Sig^{1/2}$ with power-law covariance $\Sig = \mathrm{diag}(1, 2^{-s}, \ldots, V^{-s})$~\citep{companion2026wall}, we showed that even though the nonconvex loss landscape remains free of spurious local minima, division-normalised first-order optimizers stall abruptly above a critical spectral exponent $\sstar$. At the ground truth, the smallest non-gauge Hessian eigenvalue along the rarest row $j = V$ scales at leading order as $\lambda_{\min}^{\mathrm{ng}} \propto V^{-(s+1)}D^{-s}$. Once this curvature drops below an effective denominator floor $\varepsilon_{\mathrm{eff}}$, the denominator $\sqrt{\hat{v}_t} + \varepsilon_0$ is pinned near $\varepsilon_{\mathrm{eff}}$, adaptive preconditioning is lost, and the rarest mode contracts at rate $\propto \alpha\lambda_{\min}^{\mathrm{ng}}/\varepsilon_{\mathrm{eff}}$ per step, failing to recover within a fixed step budget. Balancing $\lambda_{\min}^{\mathrm{ng}}$ against $\varepsilon_{\mathrm{eff}}$ yields the precision-wall boundary
\begin{equation}
\label{eq:wall}
\sstar(V, D) = \frac{\log(1/\varepsilon_{\mathrm{eff}}) - \log V}{\log(VD)} + \delta_{\mathrm{Adam}},
\end{equation}
where $\delta_{\mathrm{Adam}} = 0.482$ is a metric-dependent detection lag (derived and calibrated in Appendix~\ref{app:companion-wall} to keep this paper self-contained).

\begin{figure}[t]
\centering
\includegraphics[width=\linewidth]{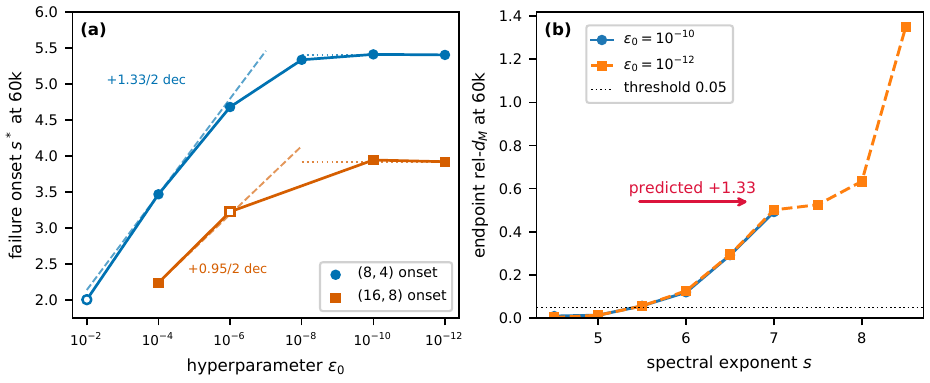}
\caption{\textbf{(a)} Empirical failure onset $\sstar$ at $60\text{k}$ steps vs.\ Adam's hyperparameter $\varepsilon_0$ across six values of $\varepsilon_0$ two decades apart ($\numEpsSweepHi$ to $\numEpsSweepLo$) in $\mathrm{fp64}$ ($n=5$ seeds, evaluated with the seed-mean linear-interpolation onset estimator of \S\ref{sec:setup}): $(V,D)=(8,4)$ across all $\numEpsSweepCount$ values ($\sstar = \numOnsetEpsTwo, \numOnsetEpsFour, \numOnsetEpsSix, \numOnsetEpsEight, \numOnsetEpsTen, \numOnsetEpsTwelve$) and $(16,8)$ at $10^{-4}, 10^{-6}, 10^{-10}, 10^{-12}$ ($\sstar = \numOnsetVOneSixDEightFour, \numOnsetVOneSixDEightSix, \numOnsetVOneSixDEightTen, \numOnsetVOneSixDEight$), shown with the $\varepsilon_0$-law reference slopes ($+\numSatMovePredicted$ and $+\numSatMovePredictedB$ per two decades, anchored at $10^{-4}$ with extrapolations $\numPredEpsTwoExtrap$ at $(8,4), 10^{-2}$ and $\numPredVOneSixDEightSixExtrap$ at $(16,8), 10^{-6}$) and saturated plateaus. Filled markers are the original sweeps; the two open markers ($(8,4)$ at $10^{-2}$; $(16,8)$ at $10^{-6}$, on a grid extended beyond the original $s \le \numCensorVOneSixDEightSixGridMax$) are re-runs with the released code, which evaluates the Adam step in a different floating-point order (\S\ref{sec:fp}; Appendix~\ref{app:ledger}); this panel shows that $\sstar$ tracks the $\varepsilon_0$ slope from $10^{-2}$ to $10^{-6}$, bends near $10^{-8}$, and saturates below $10^{-8}$. \textbf{(b)} Endpoint relative error $\relDM$ at $60\text{k}$ steps vs.\ spectral exponent $s$ at $(V,D)=(8,4)$ in re-runs with the released code for $\varepsilon_0 = 10^{-10}$ vs.\ $10^{-12}$ across matched problem instances ($n=5$ seeds), showing that the two curves superimpose at threshold crossing rather than shifting by the predicted $+\numSatMovePredicted$.}
\label{fig:eps0}
\end{figure}

Equation~\eqref{eq:wall} makes a direct causal prediction: if $\varepsilon_{\mathrm{eff}}$ is set by the hyperparameter $\varepsilon_0$, then lowering $\varepsilon_0$ by two orders of magnitude from $10^{-10}$ to $10^{-12}$ in 64-bit floating-point arithmetic ($\mathrm{fp64}$) should shift the failure boundary upward by
\[
\Delta\sstar_{\mathrm{pred}} = \frac{2\ln 10}{\ln(VD)} = \begin{cases}
\numSatMovePredicted & \text{at } (V,D) = (8,4)\ (VD = 32), \\
\numSatMovePredictedB & \text{at } (V,D) = (16,8)\ (VD = 128).
\end{cases}
\]
Instead, sweeping $\numEpsSweepCount$ values of $\varepsilon_0$ two decades apart, $\numEpsSweepHi$ to $\numEpsSweepLo$, reveals a sharp two-regime curve (Figure~\ref{fig:eps0}): $\sstar$ rises with $1/\varepsilon_0$ from $10^{-2}$ to $10^{-6}$ (at $(8,4)$, rising from $\numOnsetEpsTwo$ at $10^{-2}$---against the $\varepsilon_0$-law extrapolation $\numPredEpsTwoExtrap$ from $10^{-4}$---to $\numOnsetEpsFour$ at $10^{-4}$ and $\numOnsetEpsSix$ at $10^{-6}$, moving by $+\numEpsBranchSlope$ across the two-decade step $10^{-4}\to 10^{-6}$ and retaining $\numLadderFracFourSix\%$ of the predicted $+\numSatMovePredicted$; at $(16,8)$, rising from $\numOnsetVOneSixDEightFour$ at $10^{-4}$ to $\numOnsetVOneSixDEightSix$ at $10^{-6}$ against the $\varepsilon_0$-law extrapolation $\numPredVOneSixDEightSixExtrap$, after extending the initial $s \le \numCensorVOneSixDEightSixGridMax$ grid), retains $\numLadderFracSixEight\%$ of its slope across $10^{-6}\to 10^{-8}$, and is flat below $10^{-8}$---moving by only $\numLadderFracEightTen\%$ of the predicted step across $10^{-8}\to 10^{-10}$ and by $\numSatMoveMeasured$ at $(8,4)$ ($\numLadderFracTenTwelve\%$, $|\Delta\sstar| = \numSatMoveMeasuredAbs$) and $\numSatMoveMeasuredB$ at $(16,8)$ (from $\numOnsetVOneSixDEightTen$ to $\numOnsetVOneSixDEight$) across $10^{-10}\to 10^{-12}$.

Why does lowering $\varepsilon_0$ below $10^{-8}$ stop moving the failure boundary?

\textbf{Main claim.}
Adam generates its own internal denominator floor $\varepsilon_{\mathrm{dyn}} \propto \alpha$ whenever high-curvature and nearly flat directions are coupled by the objective. High-curvature (common) modes sit far above the adaptive edge of stability and oscillate persistently with amplitude proportional to $\alpha(1-\beta_1)/(1+\beta_1)$. Because the quartic sensing objective couples all rows of $U$, this fast-mode oscillation leaks gradient fluctuations of scale $\kappa\alpha$ into the rarest measurement row $j = V$. Those leaked fluctuations feed the rare row's second-moment accumulator $v_V$, pinning $\sqrt{\hat{v}_V} \approx \kappa\alpha$ orders of magnitude above a small hyperparameter $\varepsilon_0 \le 10^{-10}$. As a result, the effective floor in~\eqref{eq:wall} is governed by
\begin{equation}
\label{eq:eps-eff}
\varepsilon_{\mathrm{eff}}(\varepsilon_0, \alpha) \;\approx\; \max\bigl(\varepsilon_0,\, \varepsilon_{\mathrm{dyn}}(\alpha)\bigr), \qquad \varepsilon_{\mathrm{dyn}}(\alpha) \propto \alpha,
\end{equation}
and default $\varepsilon_0 = 10^{-8}$ at learning rate $\alpha = 10^{-2}$ sits directly at the crossover knee where the external hyperparameter hands control over to Adam's internal dynamical floor. Equation~\eqref{eq:eps-eff} commits to the linear scaling $\varepsilon_{\mathrm{dyn}}(\alpha) \propto \alpha$ and the monotone direction in $\beta_1$ (tested in \S\ref{sec:interventions}, items 2, 3, and 6), with its prefactor set by the order of magnitude of $\kappa$ rather than equaling any single cell's $\kappa$: specifically, $\varepsilon_{\mathrm{dyn}}$ in~\eqref{eq:wall} is a cross-grid wall-equivalent floor fitted once at \emph{threshold crossing} (onset of stall; \S\ref{sec:scale-transfer}), whereas $\kappa = \sqrt{\hat{v}_V}/\alpha$ is the \emph{post-stall} steady-state denominator measured inside the stalled regime (at $s \approx \sstar + 0.5$), which continues to vary with $s$ and $(V,D)$ after stall so that substituting post-stall $\kappa(V,D)\alpha$ back into~\eqref{eq:wall} would overpredict the $(16,8)$ plateau shift by $+\numKapCfgPlateauShift$ (at $(8,4)$, the threshold-crossing and post-stall observables differ by $\numKappaGap\times$, ratio $\numEpsDynBorrowedOverKappaAlpha$). Both numbers inherit the detection lag $\delta_{\mathrm{Adam}} = 0.482$ that the fit borrows from the companion calibration; with the lag calibrated on this paper's own $\varepsilon_0$-dominated rungs ($\delta = \numDeltaBranch$), the same fit gives $\varepsilon_{\mathrm{dyn}} = \numEpsDynBranchOverKappaAlpha\,\kappa\alpha$ (\S\ref{sec:validation}).

\textbf{Summary of evidence and scope.}
\begin{enumerate}
    \item \textbf{Elimination of non-dynamical explanations (\S\ref{sec:elimination}).} Before modeling the dynamics, we test and rule out four simpler explanations for the $\varepsilon_0 \le 10^{-8}$ plateau: (i) $\mathrm{fp32}$ underflow (all sweeps run in $\mathrm{fp64}$ with $\epsilon_{\mathrm{mach}} \approx 1.1\times 10^{-16}$); (ii) spurious local minima or non-identifiability (initializing L-BFGS and damped Gauss--Newton from Adam's stalled $60\text{k}$-step checkpoints drives relative recovery error $\relDM := \|UU^\top - M^\ast\|_F / \|M^\ast\|_F$ from $\numAdamResidLo\text{--}\numAdamResidHi$ down to $\le \numLbfgsRelMax\times 10^{-10}$ on $\numLbfgsConverged/\numLbfgsN$ L-BFGS runs and $12/\numLbfgsN$ Gauss--Newton runs); (iii) floating-point roundoff in the gradient (ULP perturbation tests confirm signal-to-roundoff ratios $> 10^8$); and (iv) a $60\text{k}$-step budget artifact (extending runs to $200\text{k}$ and $600\text{k}$ steps across $20$ seeds keeps the $\varepsilon_0 = 10^{-10}$ vs.\ $10^{-12}$ onset difference at $\numPathTenDeltaSixtyK$, $\numPathTenDeltaTwoHundredK$, and $\numPathTenDeltaSixHundredK$, ruling out an $\varepsilon_0$-controlled slow branch while documenting an $\varepsilon_0$-independent power-law creep in $T$).
    \item \textbf{Exact 1-D flip orbit, its Floquet instability, and cross-mode leakage (\S\ref{sec:mechanism}).} On a 1-D quadratic $f(x)=\frac{1}{2}\lambda x^2$, warm-started Adam has an exact period-2 flip orbit of amplitude $A(u) = A^\ast(1 - 38/u)$ with $A^\ast := \alpha/38$ for $u := \alpha\lambda/\varepsilon_0 > u_{c2} := 2(1+\beta_1)/(1-\beta_1) = 38$ (Proposition~\ref{thm:flip}), while on the Floquet-repelling burst-and-relaxation attractor the time-averaged RMS amplitude tracks $A_{\mathrm{RMS}}(u) \approx \numAuTrack\, A^\ast(1 - 38/u)$ to within $\numAuErr\%$ (where $\numAuTrack$ is the empirical least-squares burst-attractor factor over the $14$ cycling grid cells, vs.\ $\numAuErrBare\%$ against the unscaled period-2 curve; Observation~\ref{obs:flip}). We highlight upfront why the 1-D model is an entry point rather than the whole story: its transverse Floquet multiplier has modulus $|\mu|_{\max} = \numMuMid > 1$ at $(\beta_1,\beta_2)=(0.9,0.999)$ ($\numMuLo$ at $\beta_2=0.99$, $\numMuHi$ at $\beta_2=0.9999$), causing single-mode trajectories to burst and relax at rate $\beta_2^{1/2}$ (Observation~\ref{obs:flip}), and a decoupled 1-D mode loses adaptive normalisation at $\lambda_{\mathrm{stall}} = 38\varepsilon_0/\alpha \propto \varepsilon_0$. In the coupled $V \times D$ system, however, common modes stay persistently excited (sign-flip duty cycle up to $\numDutySSixZero$, RMS amplitude $\numAttrRmsLo\text{--}\numAttrRmsHi\, A^\ast$) and inject a gradient floor $\sqrt{\hat{v}_V} \approx \kappa\alpha$ (3-seed geometric mean $\kappa = \numKappa{}\times 10^{-6}$ at $(8,4)$, identical to seven significant figures between $\varepsilon_0 = 10^{-8}$ and $10^{-12}$).
    \item \textbf{Out-of-sample scale predictions and six mechanistic interventions (\S\ref{sec:validation}).} Fitting $\varepsilon_{\mathrm{dyn}} = \numEpsDyn{}\times 10^{-9}$ on five $D \ge 4$ configurations with the companion lag $\delta_{\mathrm{Adam}} = 0.482$ matches measured $\varepsilon_0 = 10^{-12}$ onsets across $V \in \{8,16,32\}$ and $D \in \{4,8\}$ with RMS $\numRmsDge$, while rank $D=2$ ($(16,2)$, residual $+\numResidDtwo$) lies outside the large-$D$ approximation. Six targeted interventions test the factors of the $\kappa\alpha \propto \alpha(1-\beta_1)/(1+\beta_1)$ mechanism (confirming the linear $\alpha$ scaling and the monotone direction in $\beta_1$, while the $\beta_1$ amplitude response is sub-linearly compressed with measured-to-predicted ratio $\numBetaOneCompZero$ at $\beta_1=0$ and $\numBetaOneCompLo\text{--}\numBetaOneCompHi$ across $\beta_1 > 0$): (i) removing cross-coordinate coupling restores the $\varepsilon_0$ slope ($\numSoftmaxSlopeMeasured$ per decade vs.\ predicted $\numSoftmaxSlopePredicted$, error $\numSoftmaxSlopeErrPct\%$); (ii) varying $\beta_1 \in \{0, 0.5, 0.8, 0.9, 0.95, 0.99\}$ confirms that both the measured floor $\sqrt{\hat{v}_V}/\alpha$ (falling monotonically from $\numRmspropRatio\times$ the $\beta_1=0.9$ level at RMSProp $\beta_1=0$ through $\numBetaOneRatioHalf, \numBetaOneRatioEight, 1.00, \numBetaOneRatioNineFive$ down to $\numBetaOneRatioNineNine\times$ at $\beta_1=0.99$) and the $60\text{k}$ onset $\sstar(10^{-12})$ (rising monotonically from $\numBetaOneWallZero$ at RMSProp $\beta_1=0$ to $\numBetaOneWallNineNine$ at $\beta_1=0.99$) remain monotone through $\beta_1 = 0.99$, while the measured-to-predicted compression ratio $[r(\beta_1)/r(0.9)]/[\frac{1-\beta_1}{1+\beta_1}/\frac{0.1}{1.9}]$ rises from $\numBetaOneCompZero$ at $\beta_1=0$ and $\numBetaOneCompLo$ at $\beta_1=0.5$ to $1.00$ at $\beta_1=0.9$ before turning back down to $\numBetaOneCompNineNine$ at $\beta_1=0.99$; (iii) lowering $\alpha$ from $10^{-2}$ to default $10^{-3}$ reduces $\sqrt{\hat{v}_V}$ by $\numDefaultLrPairRatio\times$ (median $\sqrt{\hat{v}_V}/\varepsilon_0 = \numDefaultLrRhoMedian$ at default $\varepsilon_0=10^{-8}$); (iv) weighted symmetric matrix factorization / completion exhibits the same self-generated floor ($\sqrt{\hat{v}_V}/\varepsilon_0$ median $\numFamTwoRhoMedian\times$, onset shift $\numFamTwoDelta$); (v) mini-batch sampling maps the crossover where stochastic gradient noise $\sigma_{\mathrm{mb}}$ overtakes deterministic chatter ($\numMbRhoTwelve\times$ floor amplification at $B=12$, $\numMbRhoTwentyFour\times$ at $B=24$); and (vi) operating at a smaller step size $\alpha_1$ lowers the floor directly: mid-run $10\times$ step-size decay drops fast-mode chatter duty cycle from $\numDutyPostConst$ to $\numDutyPostDecay$ and reduces stalled residuals by $\numDecaySepLo\text{--}\numDecaySepHi\times$ over continuing at $\alpha = 10^{-2}$, while comparison against an equal-budget constant-$\alpha_1$ control reveals a non-monotone trade-off across the two measured exponents, with late decay winning by $\numAlignRatioSix\times$ at $s = 6.0$ but losing by $\numAlignWorseFive\times$ ($\numAlignRatioFive\times$) at $s = 5.5$.
    \item \textbf{Scope and limitations (\S\ref{sec:discussion}).} Our theory and experiments use power-law matrix sensing and factorization where exact Hessian spectra and ground-truth recovery are available; we run no neural-network experiments, and extending the quantitative floor calibration to deep neural networks is open future work. All mathematical proofs and the self-contained companion precision-wall derivation are given in Appendix~\ref{app:proofs}, and the complete pre-registration ledger and protocol history are documented in Appendix~\ref{app:ledger}.
\end{enumerate}

\section{Problem Setup}
\label{sec:setup}

\textbf{Power-law matrix sensing.}
We recover an unknown rank-$D$ positive semidefinite matrix $M^\ast = U^\ast {U^\ast}^\top \in \R^{V \times V}$, where $U^\ast \in \R^{V \times D}$ has independent $\mathcal{N}(0, 1/V)$ entries, from $m = \lfloor 1.5VD \rfloor$ noiseless linear measurements $b_k = \langle A_k, M^\ast \rangle = \mathrm{Tr}(A_k M^\ast)$ ($k = 1, \ldots, m$). Each design matrix $A_k \in \R^{V \times V}$ is symmetrised Gaussian with power-law spectral covariance:
\begin{equation}
\label{eq:design}
A_k = \Sig^{1/2}\, \frac{G_k + G_k^\top}{2}\, \Sig^{1/2}, \qquad
\Sig = V \cdot \frac{\mathrm{diag}(1, 2^{-s}, \ldots, V^{-s})}{\sum_{j=1}^V j^{-s}}, \qquad
G_{k,ij} \stackrel{\mathrm{iid}}{\sim} \mathcal{N}\!\Bigl(0, \frac{1}{Vm}\Bigr).
\end{equation}
The trace normalisation $\mathrm{Tr}(\Sig) = V$ holds total measurement energy fixed across spectral exponents $s \ge 0$. At $s = 0$, $\Sig = I_V$ recovers isotropic sensing; as $s$ grows, measurement energy concentrates on low-index (common) rows while row $j = V$ (the rarest direction) receives weight $\Sigma_{VV} = V \cdot V^{-s} / \sum_{j=1}^V j^{-s} \sim V^{1-s}$.

\textbf{Burer--Monteiro objective and Hessian scaling convention.}
We minimise the factored mean-squared measurement error over $U \in \R^{V \times D}$:
\begin{equation}
\label{eq:objective}
L(U) = \frac{1}{m} \sum_{k=1}^m \bigl(\langle A_k, U U^\top \rangle - b_k\bigr)^2, \qquad
\nabla L(U) = \frac{4}{m} \sum_{k=1}^m \bigl(\langle A_k, U U^\top \rangle - b_k\bigr) A_k U.
\end{equation}
Note that~\eqref{eq:objective} omits the conventional $\frac{1}{4}$ prefactor sometimes used in matrix sensing, so $\nabla L(U)$ carries a factor of $4$ and the Gauss--Newton Hessian at $U = U^\ast$ carries a factor of $8/m$:
\[
H_s\bigl|_{U^\ast} = \frac{8}{m} \sum_{k=1}^m \mathrm{vec}(A_k U^\ast)\,\mathrm{vec}(A_k U^\ast)^\top, \qquad
\E\bigl[H_s \mid U^\ast\bigr] = \frac{4}{Vm}\bigl(\Sig \otimes N^\ast + \mathcal{T}\bigr),
\]
where $N^\ast := {U^\ast}^\top \Sig U^\ast \in \R^{D \times D}$ and $\mathcal{T}(\delta U) := \Sig U^\ast (\delta U)^\top \Sig U^\ast$ is the rank-$D^2$ exchange operator (derived in Appendix~\ref{app:companion-wall}). Because $UU^\top$ is invariant under $U \mapsto UQ$ for $Q \in \mathrm{O}(D)$, $H_s|_{U^\ast}$ has $D(D-1)/2$ exact gauge zero eigenvalues on the tangent space $\delta U = U^\ast \Omega$ ($\Omega^\top = -\Omega$). Transverse to the gauge orbit, the smallest non-gauge eigenvalue scales at leading order as $\lambda_{\min}^{\mathrm{ng}} \propto V^{-(s+1)}D^{-s}$.

\textbf{Adam update and failure onset definition.}
Starting from $U_0 \sim \mathcal{N}(0, 0.01\, I)$, we run Adam~\citep{kingma2015adam} in $\mathrm{fp64}$ with step size $\alpha = 10^{-2}$ (also testing $\alpha = 10^{-3}$ in \S\ref{sec:validation}), momentum parameters $(\beta_1, \beta_2) = (0.9, 0.999)$, and denominator hyperparameter $\varepsilon_0$ across six values two decades apart, $10^{-2}$ to $10^{-12}$:
\begin{equation}
\label{eq:adam}
\begin{split}
m_t &= \beta_1 m_{t-1} + (1-\beta_1) g_t, \qquad
v_t = \beta_2 v_{t-1} + (1-\beta_2) g_t^{\odot 2}, \\
\hat{m}_t &= \frac{m_t}{1-\beta_1^t}, \qquad
\hat{v}_t = \frac{v_t}{1-\beta_2^t}, \qquad
U_t = U_{t-1} - \alpha \frac{\hat{m}_t}{\sqrt{\hat{v}_t} + \varepsilon_0}.
\end{split}
\end{equation}
Recovery is measured by the gauge-invariant relative error $\relDM(U) := \|UU^\top - M^\ast\|_F / \|M^\ast\|_F$, along with the rarest-row error $\|(\Delta M)_{V,:}\|_2 / \|M^\ast_{V,:}\|_2$. For a given step budget $T$ (default $T = 60\text{k}$), the empirical failure onset $\sstar$ is defined as the first exponent $s$ at which the seed-mean $\relDM(U_T)$ crosses $0.05$, obtained by linear interpolation on an $s$-grid of spacing $0.5$.

\section{Ruling Out Non-Dynamical Explanations for $\varepsilon_0$ Saturation}
\label{sec:elimination}

When an optimizer stops responding to a precision hyperparameter below $10^{-8}$, several mundane explanations must be checked before invoking a dynamical mechanism: 32-bit floating-point underflow, spurious local minima or rank deficiency in the loss landscape, floating-point roundoff in the gradient, or simply running out of iterations. We test each in turn on the saturated branch $\varepsilon_0 \in \{10^{-8}, 10^{-10}, 10^{-12}\}$.

\subsection{Floating-point format and gradient roundoff}\label{sec:fp}

All sweeps in Figure~\ref{fig:eps0} are executed entirely in IEEE-754 double precision ($\mathrm{fp64}$, unit roundoff $u_{\mathrm{fp64}} = 2^{-53} \approx 1.11\times 10^{-16}$, smallest positive normal $2.2\times 10^{-308}$), so $\varepsilon_0 = 10^{-12}$ and $\hat{v}_t \sim 10^{-24}$ are hundreds of orders of magnitude above underflow.

To verify that $\mathrm{fp64}$ rounding errors during the matmul and trace inner products in~\eqref{eq:objective} do not create a synthetic noise floor in the rarest row $j = V$, we perturb intermediate accumulations by $\pm 1$ unit in the last place (ULP) across stalled checkpoints at $(V,D) = (8,4)$, $s \in \{4, 5, 6, 7\}$. The resulting ULP perturbation in the rarest-row gradient $g_V$ has magnitude $10^{-20}\text{--}10^{-19}$, whereas the actual rarest-row gradient norm along Adam's trajectory is $10^{-10}$ to $10^{-7}$---a signal-to-roundoff ratio exceeding $10^{8}$ ($\sqrt{\hat{v}_V}$ changes by less than $10^{-8}$ relative). Arithmetic roundoff is eight orders of magnitude too small to explain a floor at $\sim 5\times 10^{-9}$. We distinguish this single-step ULP sensitivity at a fixed checkpoint from trajectory-level divergence over $60\text{k}$ iterations: because fast modes evolve on a chaotic attractor with transverse Floquet multiplier $|\mu|_{\max} > 1$ (\S\ref{sec:mechanism}), trajectory-level quantities are not bitwise reproducible across different floating-point evaluation orders or BLAS implementations. Specifically, evaluating the Adam step as $\alpha\,\bigl(\hat{m}_t/(\sqrt{\hat{v}_t} + \varepsilon_0)\bigr)$, as the released code does, rather than as $(\alpha\,\hat{m}_t)/(\sqrt{\hat{v}_t} + \varepsilon_0)$, as in the original runs (Appendix~\ref{app:ledger} gives both code lines and the gradient expression; below we call runs of the released code \emph{re-runs}) sends individual $60\text{k}$-step trajectories onto different segments of the attractor and shifts the point estimate of $\kappa$ at $(8,4), s = 6.0$ by $\numKappaStoredOverCurrent\times$ ($\numKappa\times 10^{-6}$ in the original run vs.\ $\numKappaCurrentSix\times 10^{-6}$ in the re-run). What is reproducible across evaluation orders is the macroscopic scale of the dynamical floor and its saturation below $10^{-8}$, not a single trajectory's $\sqrt{\hat{v}_V}$: across all six $(V,D)$ configurations in Table~\ref{tab:scales}, the maximum onset shift between the two associative orderings is $|\Delta\sstar| \le \numRerunMaxShift$, and the fitted $D \ge 4$ floor $\varepsilon_{\mathrm{dyn}}$ moves by only $2\%$ (ratio $\numEpsDynRerunOverStored$, from $\numEpsDyn\times 10^{-9}$ to $\numEpsDynRerun\times 10^{-9}$).

\subsection{Landscape geometry: second-order solvers warm-started from stalled Adam checkpoints}

Could the plateau around $s \approx 5.4$ at $(8,4)$ reflect a geometric barrier---spurious local minima or information-theoretic non-identifiability at finite $m = 48$? To test the landscape directly on the exact problem instances where Adam stalls, we take all $\numLbfgsN$ stalled runs at $(V,D) = (8,4)$, $\varepsilon_0 = 10^{-12}$, $s \in \{5.5, 6.0, 6.5, 7.0\}$ ($3$ seeds per $s$), where Adam terminates at $60\text{k}$ steps with large recovery error $\relDM \in [\numAdamResidLo, \numAdamResidHi]$, and warm-start two classical second-order solvers from Adam's stalled state $U_{60\text{k}}$:
\begin{itemize}
    \item \textbf{L-BFGS} (strong Wolfe line search) converges to machine precision ($\relDM \le \numLbfgsRelMax\times 10^{-10}$) on $\numLbfgsConverged$ of the $\numLbfgsN$ stalled instances within $\numLbfgsStepsLo\text{k}\text{--}\numLbfgsStepsHi\text{k}$ iterations.
    \item \textbf{Levenberg--Marquardt / damped Gauss--Newton} (solving $(H_{\mathrm{GN}} + \mu I)\delta U = -\nabla L(U)$ with adaptive damping $\mu$) converges on all $12$ of $\numLbfgsN$ stalled instances to $\relDM \le 10^{-12}$.
\end{itemize}
Because every stalled Adam checkpoint lies in a smooth basin connected directly to $M^\ast$ and solvable to $10^{-12}$ accuracy once the full $VD \times VD$ curvature is inverted, the barrier is purely a property of Adam's diagonal first-order dynamics.

\subsection{Finite step budget: decoupling $\varepsilon_0$-insensitivity from power-law budget creep}

A natural concern (raised whenever a threshold $\relDM > 0.05$ is evaluated at a fixed step count $T = 60\text{k}$) is that rare modes with tiny curvature $\lambda_{\min}^{\mathrm{ng}}$ simply need more steps to converge, or that the plateau depends on the choice of $0.05$ threshold or linear interpolation; we control for the step budget in Table~\ref{tab:budget-sweep} and for the threshold and interpolant via the threshold-free denominator measurement $\sqrt{\hat{v}_V}$ (\S\ref{sec:cross-mode}) and the log-vs.-linear interpolation audit (Appendix~\ref{app:ledger}, item 1).

To test the budget dependence directly, we extend the $(V,D) = (8,4)$ sweep from $T = 60\text{k}$ to $T = 200\text{k}$ and $T = 600\text{k}$ steps across $n = 10$ matched seeds per arm on $s \in \{5.0, 5.5, 6.0, 6.5, 7.0\}$ at $\varepsilon_0 \in \{10^{-10}, 10^{-12}\}$ (and $n = 20$ seeds at $\varepsilon_0 = 10^{-12}$, generating $\numBootNSixtyK$ five-seed subsets to measure sampling variability; Table~\ref{tab:budget-sweep}), and also run long trajectories up to $T = 10^7$ steps. At $600\text{k}$ steps, $\numBootCensoredSixHundredK$ of the $\numBootNSixHundredK$ subsets fall below the $s$-grid and are censored ($\numBootCensoredSixtyK$ at $60\text{k}$ and $\numBootCensoredTwoHundredK$ at $200\text{k}$), and the uncensored 5-of-20 subset 95\% width at $600\text{k}$ ($\numBootRangeSixHundredK$) is of the same order as the two-decade $\varepsilon_0$-law prediction $+\numSatMovePredicted$. Consequently, our test criterion is not the $n = 20$ absolute onset across independent subsets, but the matched-seed difference $\Delta\sstar(10^{-12} - 10^{-10})$, whose leave-one-seed-out (LOO) intervals, fixed before the runs (Column 5 of Table~\ref{tab:budget-sweep}), exclude $+\numSatMovePredicted$ at every budget.

\begin{table}[t]
\centering\footnotesize
\setlength{\tabcolsep}{4pt}
\caption{Multi-budget test of $\varepsilon_0$-insensitivity at $(V,D) = (8,4)$. While the absolute onset $\sstar(T)$ creeps upward with step budget $T$, the matched-seed difference $\Delta\sstar(10^{-12} - 10^{-10})$ remains near zero at every budget (against the two-decade $\varepsilon_0$-law prediction of $+\numSatMovePredicted$). Columns 2--5 use seeds 0--9 ($n=10$); the $(8,4)$ row of Table~\ref{tab:scales} uses seeds 0--4 and gives $\numOnsetVEightDFour$ in the original runs and $\numRerunVEightDFour$ in the re-runs, both inside the $60\text{k}$ 5-of-20 range below, so the two tables differ by seed set, not by estimator. Columns 6--7 report the $n=20$ onset and the 95\% range across all $\numBootNSixtyK$ five-seed subsets at $\varepsilon_0 = 10^{-12}$ (each subset onset uses the same seed-mean linear-interpolation estimator as every other onset in this paper, \S\ref{sec:setup}; at $(8,4)$, the 5-of-20 subset SD is $\numBootSdSixtyK$ at $60\text{k}$ and $\numBootSdSixHundredK$ at $600\text{k}$ with $\numBootCensoredSixHundredK$ censored subsets dropped at $600\text{k}$, Appendix~\ref{app:ledger}, item 2).}
\label{tab:budget-sweep}
\begin{tabular}{r|rr|cc|cc}
\toprule
$T$ & $\sstar(10^{-10})$ & $\sstar(10^{-12})$ & $\Delta\sstar(12{-}10)$ & LOO of $\Delta\sstar$ & $\sstar(10^{-12})$ ($n{=}20$) & 5-of-20 95\% (SD) \\
\midrule
$60\text{k}$  & \numPathTenOnsetTenSixtyK      & \numPathTenOnsetTwelveSixtyK      & $\numPathTenDeltaSixtyK$      & $[\numPathTenDeltaSixtyKLooLo,\, \numPathTenDeltaSixtyKLooHi]$      & \numOnsetTwentySixtyK      & $[\numBootLoSixtyK,\, \numBootHiSixtyK]$ (\numBootSdSixtyK) \\
$200\text{k}$ & \numPathTenOnsetTenTwoHundredK & \numPathTenOnsetTwelveTwoHundredK & $\numPathTenDeltaTwoHundredK$ & $[\numPathTenDeltaTwoHundredKLooLo,\, \numPathTenDeltaTwoHundredKLooHi]$ & \numOnsetTwentyTwoHundredK & $[\numBootLoTwoHundredK,\, \numBootHiTwoHundredK]$ (\numBootSdTwoHundredK) \\
$600\text{k}$ & \numPathTenOnsetTenSixHundredK & \numPathTenOnsetTwelveSixHundredK & $\numPathTenDeltaSixHundredK$ & $[\numPathTenDeltaSixHundredKLooLo,\, \numPathTenDeltaSixHundredKLooHi]$ & \numOnsetTwentySixHundredK & $[\numBootLoSixHundredK,\, \numBootHiSixHundredK]$ (\numBootSdSixHundredK) \\
\bottomrule
\end{tabular}
\end{table}

Table~\ref{tab:budget-sweep} separates two distinct phenomena that must not be conflated:
\begin{enumerate}
    \item \textbf{Within-budget $\varepsilon_0$-insensitivity holds at all budgets.} At $T = 60\text{k}$, $200\text{k}$, and $600\text{k}$, the matched-seed onset difference between $\varepsilon_0 = 10^{-10}$ and $10^{-12}$ is $\numPathTenDeltaSixtyK$, $\numPathTenDeltaTwoHundredK$, and $\numPathTenDeltaSixHundredK$ (Table~\ref{tab:budget-sweep}; the $n=5$ comparison at $60\text{k}$ is in Table~\ref{tab:dispersion}), ruling out the two-decade $\varepsilon_0$-law prediction $+\numSatMovePredicted$ across all three budgets.
    \item \textbf{$\varepsilon_0$-independent power-law creep in $T$.} Across $T \in [60\text{k}, 10^7]$, the stalled residual at fixed $s > \sstar$ decays slowly as a power law in $T$ (with aggregate-to-rarest creep rate ratio $\numCreepRatio$ and $\numCreepRatioRare$), causing the $0.05$-threshold crossing $\sstar(T)$ to drift upward from $\numPathTenOnsetTwelveSixtyK$ at $60\text{k}$ to $\numPathTenOnsetTwelveSixHundredK$ at $600\text{k}$ (and $\numPNineOnsetLog$ under log-interpolation at $600\text{k}$; Table~\ref{tab:dispersion}). Crucially, this slow creep occurs at the \emph{same} rate for $\varepsilon_0 = 10^{-8}$, $10^{-10}$, and $10^{-12}$ because the denominator $\sqrt{\hat{v}_V} \approx \kappa\alpha$ is pinned independently of $\varepsilon_0$, turning rare-mode evolution into an effective step-size $\alpha_{\mathrm{eff},V} \sim \alpha/(\kappa\alpha) = 1/\kappa$ gradient flow with tiny curvature $\lambda_{\min}^{\mathrm{ng}}$. Thus the "precision wall" at $60\text{k}$ steps is a sharp finite-budget crossover into an $\varepsilon_0$-independent slow-creep regime rather than a $T = \infty$ asymptote.
\end{enumerate}

\section{Mechanism: Fast-Mode Flip Orbits and Cross-Mode Gradient Leakage}
\label{sec:mechanism}

To understand why $\sqrt{\hat{v}_V}$ is pinned near $\kappa\alpha \gg \varepsilon_0$, we proceed in two steps. First (\S\ref{sec:1d-orbit}), we analyse a single quadratic mode under warm-started Adam to establish the exact period-2 flip orbit amplitude and the dimensionless stability parameter $u := \alpha\lambda/\varepsilon_0$. We also state explicitly why the 1-D model cannot by itself produce an $\varepsilon_0$-independent floor. Second (\S\ref{sec:cross-mode}), we show how in the coupled $V \times D$ system, persistent fast-mode oscillation leaks gradient fluctuations of scale $\kappa\alpha$ into the rarest row.

\subsection{Single-mode period-2 flip orbit and its two structural limits}
\label{sec:1d-orbit}

Consider warm-started Adam ($\hat{m}_t = m_t$, $\hat{v}_t = v_t$) with step size $\alpha > 0$, momentum parameters $\beta_1 \in [0,1)$, $\beta_2 \in (0,1)$, and denominator constant $\varepsilon_0 > 0$ on a 1-D quadratic $f(x) = \frac{1}{2}\lambda x^2$ ($g_t = \lambda x_t$). (In our matrix-sensing experiments, Adam uses standard bias correction $\hat{m}_t = m_t/(1-\beta_1^t)$ and $\hat{v}_t = v_t/(1-\beta_2^t)$ as in~\eqref{eq:adam}; because the correction factors differ negligibly from $1$ once $t \gg 1/(1-\beta_2) = 10^3$ steps---well before the $60\text{k}$-step budget---the warm-started recurrence below describes the post-transient orbit.) The following proposition characterizes its exact period-2 alternating orbit and linear stability.

\begin{proposition}[1-D period-2 flip orbit and Floquet stability]
\label{thm:flip}
Define the dimensionless curvature-to-precision ratio $u := \alpha\lambda/\varepsilon_0$, the adaptive edge-of-stability threshold $u_{c2} := \frac{2(1+\beta_1)}{1-\beta_1}$ ($u_{c2} = 38$ at $\beta_1 = 0.9$), and the limiting flip amplitude $A^\ast := \frac{\alpha(1-\beta_1)}{2(1+\beta_1)} = \frac{\alpha}{u_{c2}}$. Then:
\begin{enumerate}
    \item \textbf{Existence and exact amplitude.} A positive-amplitude period-2 orbit $x_{t+1} = -x_t$ with $|x_t| = A > 0$, $m_t = -m_{t-1}$, and stationary $v_t = (\lambda A)^2$ exists if and only if $u > u_{c2}$, equivalently
    \begin{equation}
    \label{eq:lambda-stall}
    \lambda > \lambda_{\mathrm{stall}} := \frac{u_{c2}\,\varepsilon_0}{\alpha} = \frac{2(1+\beta_1)}{1-\beta_1}\frac{\varepsilon_0}{\alpha},
    \end{equation}
    and its unique amplitude is
    \begin{equation}
    \label{eq:au}
    A(u) = A^\ast\Bigl(1 - \frac{u_{c2}}{u}\Bigr) = \frac{\alpha(1-\beta_1)}{2(1+\beta_1)} - \frac{\varepsilon_0}{\lambda}.
    \end{equation}
    \item \textbf{Contraction below $\lambda_{\mathrm{stall}}$.} For $0 < u < u_{c2}$, the origin $x = 0$ is locally asymptotically stable: convergence to $x=0$ is monotone for $0 < u \le u_{c1} := \frac{(1-\sqrt{\beta_1})^2}{1-\beta_1}$ ($u_{c1} \approx 0.0263$ at $\beta_1 = 0.9$) and damped oscillatory for $u_{c1} < u < u_{c2}$.
    \item \textbf{Transverse Floquet instability for $\beta_1 > 0$.} Across a $\numFloqGridN$-cell $(\beta_1, \beta_2)$ grid, the two-step return map $F^2$ linearized around the period-2 orbit repels on $\numFloqRepelCount$ cells, with stability boundary crossing $|\mu|_{\max} = 1$ at $\beta_1 \in [\numFloqCrossLo, \numFloqCrossHi]$. Specifically, for $\beta_1 > \numFloqCrossHi$ and $u > u_{c2}$, the transverse $(x,m)$ momentum block has spectral radius $|\mu|_{\max} > 1$ (at $\beta_1 = 0.9$, $|\mu|_{\max} = \numMuLo, \numMuMid, \numMuHi$ for $\beta_2 = 0.99, 0.999, 0.9999$), whereas at $\beta_1 = 0$ (RMSProp) the momentum state disappears and $|\mu|_{\max} = \numFloqZeroMuMid = \beta_2 < 1$.
\end{enumerate}
\end{proposition}

The proof is given in Appendix~\ref{app:proof-flip}. Proposition~\ref{thm:flip} establishes both what the 1-D model explains and the two exact reasons why a 1-D model cannot by itself explain the $\varepsilon_0 \le 10^{-8}$ saturation in matrix sensing:
\begin{enumerate}
    \item \textbf{In 1-D, the threshold $\lambda_{\mathrm{stall}} = 38\varepsilon_0/\alpha$ still scales linearly with $\varepsilon_0$.} If the rarest mode evolved as an isolated 1-D quadratic, lowering $\varepsilon_0$ by $10^4$ would lower $\lambda_{\mathrm{stall}}$ by $10^4$ and shift $\sstar$ upward by $4\ln 10/\ln(VD)$.
    \item \textbf{In 1-D, the period-2 orbit is Floquet-repelling for $\beta_1 = 0.9$.} Because $|\mu|_{\max} = \numMuMid > 1$, a generic 1-D trajectory does not stay glued to $x_{t+1} = -x_t$; instead, phase drift between $x_t$ and $m_t$ periodically cancels the step and sends the 1-D trajectory into a two-stage decay governed by $\beta_2^{1/2}$, as summarized next.
\end{enumerate}

\begin{figure}[t]
\centering
\includegraphics[width=0.62\linewidth]{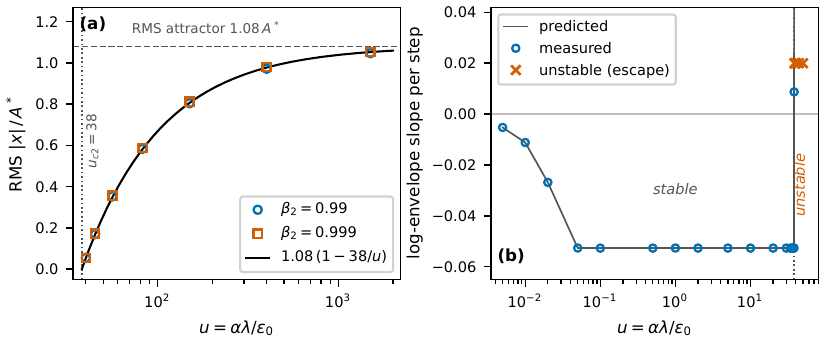}
\caption{\textbf{(a)} Verification of the amplitude law~\eqref{eq:au} across $(\varepsilon_0, \alpha, \lambda)$ spanning $\numAuConjEpsSpan\times$ in $\varepsilon_0$ and $\numAuConjAlphaSpan\times$ in $\alpha$: rescaled time-averaged RMS amplitude $A_{\mathrm{RMS}}/A^\ast$ on the burst-and-relaxation attractor collapses onto the empirical tracking curve $\numAuTrack\,(1 - 38/u)$ across $\beta_2 \in \{0.99, 0.999\}$ (within $\numAuErr\%$, vs.\ $\numAuErrBare\%$ against the unscaled exact period-2 orbit $1 - 38/u$; worst cross-leg spread $\numAuConjSpread\%$ at matched $u$; Observation~\ref{obs:flip}). \textbf{(b)} Per-step log-envelope slope of the flip amplitude vs.\ $u = \alpha\lambda/\varepsilon_0$ (stable below, unstable/escaping above the stability threshold $u_{c2} = 38$).}
\label{fig:theory}
\end{figure}

\begin{observation}[1-D two-stage burst-and-relaxation cycle]
\label{obs:flip}
Across the $\numCycleRows$ grid cells classified as sustained cycling ($u > u_{c2}$), transverse instability drives recurrent flares (near-period-2 carrier period $\numAttrPeriodLo\text{--}\numAttrPeriodHi$ steps, coarse-to-fine crossing alias ratio $\numAliasLo\text{--}\numAliasHi\times$): when $x_t$ and $m_t$ dephase, $v_t$ decays exponentially as $v_t \propto \beta_2^t$ so $\sqrt{v_t}$ relaxes at rate $\beta_2^{1/2}$ per step, letting $|x_t|$ contract until the effective step size $\alpha\lambda/(\sqrt{v_t} + \varepsilon_0)$ re-crosses $u_{c2}$ and triggers a new burst (with $\numExtWithin\%$ of late-window extrema of $\sqrt{\hat{v}_t}/(\lambda A^\ast)$ lying within $\pm 25\%$ of $1$, median extrema $[\numExtMedLo, \numExtMedHi]$, and $\numRowsInBand\%$ of rows keeping their entire late window within $\pm 25\%$; right at the boundary $u = \numBoundaryU$, late-window $\sqrt{\hat{v}_t}/(\lambda A^\ast)$ drops to $[\numBoundaryLo, \numBoundaryHi]$). Time-averaging over sustained cycles yields an empirical RMS amplitude tracking $A_{\mathrm{RMS}}(u) \approx \numAuTrack\, A^\ast(1 - 38/u)$ to within $\numAuErr\%$ ($\numAuErrBare\%$ against the unscaled $A^\ast(1 - 38/u)$ curve; Figure~\ref{fig:theory}).
\end{observation}

\subsection{Cross-mode gradient leakage pins $\sqrt{\hat{v}_V} \approx \kappa\alpha$ in the coupled system}
\label{sec:cross-mode}

Why do fast modes remain persistently active in the $V \times D$ Burer--Monteiro problem instead of relaxing to zero, and how do they pin the rarest row's denominator?

In the full sensing objective $L(U) = \frac{1}{m}\sum_{k=1}^m (\langle A_k, UU^\top\rangle - b_k)^2$, the measurement residuals $r_k(U) = \langle A_k, UU^\top - M^\ast\rangle$ are shared across all rows $i \in \{1, \ldots, V\}$. Specifically, the gradient on row $i$ is
\begin{equation}
\label{eq:coupled-grad}
g_i(U) = (\nabla L(U))_{i,:} = \frac{4}{m} \sum_{k=1}^m r_k(U)\, (A_k U)_{i,:}.
\end{equation}
Let $U = U^\ast + \delta U$, where common rows ($i \ll V$, large curvature $\lambda_i \gg \lambda_{\mathrm{stall}}$, $u_i \sim 10^6\text{--}10^{10}$) oscillate at the edge of stability with displacement $\|\delta U_{\mathrm{fast}}\| \sim A^\ast = \frac{\alpha(1-\beta_1)}{2(1+\beta_1)}$. Because the $O(VD)$ high-curvature modes have incommensurate curvatures, their flare phases do not synchronize simultaneously: while one fast mode briefly dephases, others are near peak burst, keeping the shared residual $r_k(U) \propto \|\delta U_{\mathrm{fast}}\| \propto \alpha\frac{1-\beta_1}{1+\beta_1}$ nonzero on a high fraction of steps (measured fast-mode chatter duty cycle rising from $\numDutySFourZero$ and $\numDutySFourFive$ at $s=4.0, 4.5$ to $\numDutySFiveZero, \numDutySFiveFive, \numDutySSixZero$ at $s=5.0, 5.5, 6.0$, with RMS amplitude $\numAttrRmsLo\text{--}\numAttrRmsHi\, A^\ast$ and $\numRmsAmp\, A^\ast$ on sustained cycles; this multi-mode persistence is a numerical finding supported by the duty-cycle measurements in Figure~\ref{fig:decay} and phase-incommensurability arguments rather than a formal multi-mode Floquet theorem, \S\ref{sec:discussion}, item 2).

Through the off-diagonal entries of $A_k = \Sig^{1/2} G_{\mathrm{sym},k} \Sig^{1/2}$ in~\eqref{eq:coupled-grad}, these persistent fast-mode residuals $r_k(U)$ drive a fluctuating gradient component on the rarest row $i = V$ even when row $V$'s own restoring curvature $\lambda_V \propto V^{-(s+1)}D^{-s}$ is negligible:
\begin{equation}
\label{eq:leakage}
\|g_V(t)\|_{\mathrm{leak}} \;\sim\; \kappa\,\alpha, \qquad \kappa \;\propto\; \frac{1-\beta_1}{1+\beta_1}.
\end{equation}
Because Adam's second-moment accumulator averages squared gradients over a $(1-\beta_2)^{-1} = 10^3$-step window, this leaked gradient variance pins the rarest row's denominator at
\begin{equation}
\label{eq:vhat-floor}
\sqrt{\hat{v}_V} \;\approx\; \kappa\,\alpha.
\end{equation}

\textbf{Direct measurement of $\sqrt{\hat{v}_V}/\alpha$ across four decades of $\varepsilon_0$.}
Measuring the stationary rare-row denominator $\sqrt{\hat{v}_V}$ directly on stalled trajectories at $(V,D) = (8,4)$ ($s = 6.0$, $\alpha = 10^{-2}$) confirms~\eqref{eq:vhat-floor}:
\begin{itemize}
    \item At $\varepsilon_0 = 10^{-8}$, $10^{-10}$, and $10^{-12}$, the 3-seed geometric mean of $\sqrt{\hat{v}_V}/\alpha$ is $\numKappa \times 10^{-6}$ (so $\sqrt{\hat{v}_V} = \numSelfEps \times 10^{-8}$ at $\alpha = 10^{-2}$).
    \item For each individual seed, changing $\varepsilon_0$ by four orders of magnitude from $10^{-8}$ to $10^{-12}$ leaves $\sqrt{\hat{v}_V}/\alpha$ identical to \textbf{seven significant figures}; across the three seeds at $s = 6.0$, the per-seed values are $5.647\times 10^{-6}$, $5.650\times 10^{-6}$, and $5.657\times 10^{-6}$ (within $0.2\%$ of one another), while across all $\numKappaSCells$ stalled exponent cells $s \in [4.5, 7.5]$ at $(8,4)$ the cell-mean varies by a factor of $\numKappaSSpread\times$ (and at $(16,8)$ the 10-pair geometric mean is $\numKapCfgSixteenEight \times 10^{-6}$, ratio $\numKapCfgRatio$ to $(8,4)$).
\end{itemize}
Because $\sqrt{\hat{v}_V} \approx \numSelfEps \times 10^{-8}$ is already five times larger than $\varepsilon_0 = 10^{-8}$ and five thousand times larger than $\varepsilon_0 = 10^{-12}$, adding $\varepsilon_0 \le 10^{-10}$ to $\sqrt{\hat{v}_V}$ in the denominator $\sqrt{\hat{v}_V} + \varepsilon_0$ has virtually no effect on the update.

We emphasize that $\varepsilon_{\mathrm{dyn}}$ and $\kappa$ are two distinct observables: $\varepsilon_{\mathrm{dyn}}$ in~\eqref{eq:wall} is a cross-grid wall-equivalent floor fitted once at \emph{threshold crossing} (onset of stall; \S\ref{sec:scale-transfer}), whereas $\kappa = \sqrt{\hat{v}_V}/\alpha$ is the \emph{post-stall} steady-state denominator measured inside the stalled regime (at $s \approx \sstar + 0.5$), which continues to vary with $s$ and $(V,D)$ after stall so that post-stall $\kappa(V,D)\alpha$ cannot be substituted back into the threshold formula~\eqref{eq:sstar-dyn}. What~\eqref{eq:eps-eff}--\eqref{eq:vhat-floor} commit to, and what is directly verified in \S\ref{sec:interventions}, is the linear scaling $\sqrt{\hat{v}_V} \propto \alpha$ (item 3, $\numDefaultLrPairRatio\times$, and item 6) and the monotone direction in $\beta_1$ (item 2), rather than the cell-by-cell prefactor linking post-stall $\kappa$ to threshold-crossing $\varepsilon_{\mathrm{dyn}}$. Substituting the fitted threshold-crossing floor $\varepsilon_{\mathrm{dyn}} = \numEpsDyn \times 10^{-9}$ ($\approx \numEpsDynBorrowedOverKappaAlpha\,\kappa\alpha$, a factor of $\numKappaGap\times$ below post-stall $\kappa\alpha$ at $(8,4)$, under the companion lag $\delta_{\mathrm{Adam}} = 0.482$; $\approx \numEpsDynBranchOverKappaAlpha\,\kappa\alpha$ under the lag $\delta = \numDeltaBranch$ calibrated on this paper's own $\varepsilon_0$-dominated rungs, \S\ref{sec:validation}; Table~\ref{tab:dispersion} lists the walls an unscaled $\kappa\alpha$ would give) into~\eqref{eq:eps-eff} explains the $\varepsilon_0$ curve across six values of $\varepsilon_0$ two decades apart ($10^{-2}$ to $10^{-12}$) in Figure~\ref{fig:eps0}(a):
\begin{itemize}
    \item For $\varepsilon_0 \ge 10^{-6}$, $\varepsilon_0 \gg \varepsilon_{\mathrm{dyn}}$, so $\varepsilon_{\mathrm{eff}} \approx \varepsilon_0$ and $\sstar$ follows the hyperparameter slope $\ln(1/\varepsilon_0)/\ln(VD)$.
    \item Across $\varepsilon_0 \in [10^{-6}, 10^{-8}]$, $\varepsilon_0$ approaches $\sqrt{\hat{v}_V} \approx 5.65\times 10^{-8}$ and $\varepsilon_{\mathrm{dyn}} \approx 4.76\times 10^{-9}$, reducing the two-decade slope to $\numLadderFracSixEight\%$ of its asymptotic value.
    \item For $\varepsilon_0 \le 10^{-8}$, $\varepsilon_{\mathrm{dyn}}$ dominates and $\sstar$ saturates at the horizontal plateau
    \begin{equation}
    \label{eq:sstar-dyn}
    \sstar_{\mathrm{plateau}}(V, D) = \frac{\log(1/\varepsilon_{\mathrm{dyn}}) - \log V}{\log(VD)} + \delta_{\mathrm{Adam}}.
    \end{equation}
\end{itemize}

\section{Empirical Validation and Mechanistic Interventions}
\label{sec:validation}

We test~\eqref{eq:sstar-dyn} and the cross-mode leakage mechanism~\eqref{eq:leakage}--\eqref{eq:vhat-floor} in two ways: first (\S\ref{sec:scale-transfer}), by checking whether a single $\varepsilon_{\mathrm{dyn}}$ fitted with frozen companion lag $\delta_{\mathrm{Adam}} = 0.482$ predicts the saturated plateau onset across dimensions $(V,D)$; second (\S\ref{sec:interventions}), by intervening on each physical ingredient of the mechanism---cross-mode coupling, $\beta_1$, step size $\alpha$, matrix factorization family, mini-batch noise, and step-size decay.

\subsection{Multi-scale $(V,D)$ validation on the saturated $\varepsilon_0 = 10^{-12}$ branch}
\label{sec:scale-transfer}

With the detection lag frozen at its companion calibration value $\delta_{\mathrm{Adam}} = 0.482$ (Appendix~\ref{app:companion-wall}), fitting the single wall-equivalent scalar $\varepsilon_{\mathrm{dyn}}$ to the five $D \ge 4$ configurations at $\varepsilon_0 = 10^{-12}$ gives $\varepsilon_{\mathrm{dyn}} = \numEpsDyn \times 10^{-9}$ ($\numEpsDynLog \times 10^{-9}$, a $\numEpsDynLogPct\%$ shift, under log-interpolation); as shown below, $\varepsilon_{\mathrm{dyn}}$ is an empirical cross-grid wall parameter rather than a cell-by-cell algebraic function of post-stall $\kappa(V,D)\alpha$. Table~\ref{tab:scales} compares~\eqref{eq:sstar-dyn} across all six configurations spanning $V \in \{8, 16, 32\}$ and $D \in \{2, 4, 8\}$.

\begin{table}[t]
\centering\footnotesize
\setlength{\tabcolsep}{3.5pt}
\caption{Multi-scale validation of the saturated plateau prediction~\eqref{eq:sstar-dyn} at $\varepsilon_0 = 10^{-12}$ ($n=5$ seeds, $60\text{k}$ steps; each onset is the seed-mean linear-interpolation onset of \S\ref{sec:setup} over seeds 0--4, and at $(8,4)$ the 5-of-20 subset SD is $\numBootSdSixtyK$ at $60\text{k}$ and $\numBootSdSixHundredK$ at $600\text{k}$, Appendix~\ref{app:ledger}, item 2). Here $\delta_{\mathrm{Adam}} = 0.482$ is frozen from the companion $\varepsilon_0 = 10^{-8}$ calibration and $\varepsilon_{\mathrm{dyn}} = \numEpsDyn \times 10^{-9}$ is fitted across the five $D \ge 4$ rows of the original runs (RMS $\numRmsDge$). Column 7 reports the re-run onsets (\S\ref{sec:fp}) and their shifts $\Delta$ from Column 6, which give $\varepsilon_{\mathrm{dyn}} = \numEpsDynRerun \times 10^{-9}$ (ratio $\numEpsDynRerunOverStored$ to the original fit) and $D \ge 4$ RMS $\numRmsDgeRerun$. Column 8 reports the post-stall 5-seed geometric mean $\kappa = \sqrt{\hat{v}_V}/\alpha$ ($\times 10^{-6}$), measured in re-runs at $s \approx \sstar(10^{-12}) + 0.5$.}
\label{tab:scales}
\begin{tabular}{rr|rr|rrrr|l}
\toprule
$V$ & $D$ & $VD$ & $m$ & pred.\ $\sstar_{\mathrm{plateau}}$ & $\sstar(10^{-12})$ & re-run $\sstar(10^{-12})$ ($\Delta$) & $\kappa\ (10^{-6})$ & domain \\
\midrule
8  & 4 & 32  & 48  & \numPredVEightDFour     & \numOnsetVEightDFour     & \numRerunVEightDFour\ ($\numRerunShiftVEightDFour$)       & \numKapCellVEightDFour     & $D \ge 4$ \\
16 & 4 & 64  & 96  & \numPredVOneSixDFour    & \numOnsetVOneSixDFour    & \numRerunVOneSixDFour\ ($\numRerunShiftVOneSixDFour$)     & \numKapCellVOneSixDFour    & $D \ge 4$ \\
16 & 8 & 128 & 192 & \numPredVOneSixDEight   & \numOnsetVOneSixDEight   & \numRerunVOneSixDEight\ ($\numRerunShiftVOneSixDEight$)   & \numKapCellVOneSixDEight   & $D \ge 4$ \\
32 & 4 & 128 & 192 & \numPredVThreeTwoDFour  & \numOnsetVThreeTwoDFour  & \numRerunVThreeTwoDFour\ ($\numRerunShiftVThreeTwoDFour$) & \numKapCellVThreeTwoDFour  & $D \ge 4$ \\
32 & 8 & 256 & 384 & \numPredVThreeTwoDEight & \numOnsetVThreeTwoDEight & \numRerunVThreeTwoDEight\ ($\numRerunShiftVThreeTwoDEight$) & \numKapCellVThreeTwoDEight & $D \ge 4$ \\
\midrule
16 & 2 & 32  & 48  & \numPredVOneSixDTwo     & \numOnsetVOneSixDTwo     & \numRerunVOneSixDTwo\ ($\numRerunShiftVOneSixDTwo$)       & \numKapCellVOneSixDTwo     & $D = 2$ ($\Delta = +\numResidDtwo$) \\
\bottomrule
\end{tabular}
\end{table}

Across all five $D \ge 4$ configurations in Table~\ref{tab:scales}, the saturated plateau prediction matches measured onsets with RMS $\numRmsDge$ on the original runs and $\numRmsDgeRerun$ on the re-runs ($\varepsilon_{\mathrm{dyn}} = \numEpsDynRerun\times 10^{-9}$, ratio $\numEpsDynRerunOverStored$ to the original fit; $\numRmsDgeLog$ under log-interpolation, against $\numRmsVDFree$ for a free-constant product-only $VD$ law). Each onset is a seed-mean interpolation over $n=5$ seeds, and LOO intervals on five seeds understate the variability across independent seed sets: in our 20-seed ensemble at $(8,4)$, the SD across 5-of-20 subsets is $\numBootSdSixtyK$ at $60\text{k}$, $\numBootOverLoo\times$ the five-seed LOO half-width (Appendix~\ref{app:ledger}, item 2; Table~\ref{tab:dispersion}). Moreover, at fixed $VD = 128$ ($(32,4)$ vs.\ $(16,8)$), the larger-$V$ configuration walls earlier ($\numOnsetVThreeTwoDFour$ vs.\ $\numOnsetVOneSixDEight$ in the original runs; $\numRerunVThreeTwoDFour$ vs.\ $\numRerunVOneSixDEight$ in the re-runs), matching the parameter-free $-\log 2/\log(128) = -0.143$ shift. Just as at $\varepsilon_0 = 10^{-8}$, rank $D = 2$ ($(16,2)$) walls higher than predicted ($\Delta = +\numResidDtwo$, raising the 6-cell RMS to $\numRmsAll$ with $\varepsilon_{\mathrm{dyn}} = \numEpsDynAll\times 10^{-9}$) because a two-column projection $N^\ast \in \R^{2\times 2}$ sits in the small-$D$ spectral regime (Appendix~\ref{app:companion-wall}). Because $\delta_{\mathrm{Adam}}$ and $\log(1/\varepsilon_{\mathrm{dyn}})$ enter~\eqref{eq:sstar-dyn} through $\log(1/\varepsilon_{\mathrm{dyn}})/\log(VD) + \delta_{\mathrm{Adam}}$ over a finite range $\log(VD) \in [\ln 32, \ln 256]$, jointly fitting both parameters on the five $D \ge 4$ rows yields a nearly flat ridge (RMS between $\numRmsSpanLo$ and $\numRmsSpanHi$ for $\delta \in [0.482, 0.660]$, while $\varepsilon_{\mathrm{dyn}}$ changes by $\numEpsDynRatio\times$; other fits in Table~\ref{tab:dispersion}), so the five rows alone cannot separate the two parameters, and we freeze $\delta_{\mathrm{Adam}} = 0.482$ from the independent companion calibration. That calibration uses a different observable from the one fitted here (the rarest-direction onset at $\varepsilon_0 = 10^{-8}$, $n = 10$; Appendix~\ref{app:companion-wall}), and the fitted prefactor depends on this choice: calibrating the lag instead on this paper's own $\varepsilon_0$-dominated rungs at $(8,4)$ ($\varepsilon_0 = 10^{-4}, 10^{-6}$) gives $\delta = \numDeltaBranch$, $\varepsilon_{\mathrm{dyn}} = \numEpsDynBranch\times 10^{-9}$ and RMS $\numRmsDgeBranch$, which is $\numEpsDynBranchOverKappaAlpha$ rather than $\numEpsDynBorrowedOverKappaAlpha$ in units of the measured $(8,4)$ $\kappa\alpha$. The flat plateau below $\varepsilon_0 = 10^{-8}$ and the $\alpha$ scaling do not depend on the lag; the prefactor of $\varepsilon_{\mathrm{dyn}}$ relative to $\kappa\alpha$ is fixed only to within this range.

\textbf{Wall-equivalent scalar $\varepsilon_{\mathrm{dyn}}$ vs.\ configuration-dependent $\kappa(V,D,s)$.}
Why does a single constant scalar $\varepsilon_{\mathrm{dyn}}$ fit all five $D \ge 4$ rows in Table~\ref{tab:scales} even though the expected off-diagonal gradient coupling scales as $\kappa(V,D,s) \propto V^{-(s+1)/2}D^{-1}$ (Appendix~\ref{app:companion-wall}, item 4)? In an initial two-configuration comparison, the measured $\kappa = \sqrt{\hat{v}_V}/\alpha$ dropped from $\numKappaCurrentSix\times 10^{-6}$ at $(8,4), s = 6.0$ (re-run) to $\numKapCfgSixteenEight\times 10^{-6}$ across $s \in \{4.5, 5.0\}$ at $(16,8)$---a ratio of $\numKapCfgRatio\times$ that, if substituted directly into~\eqref{eq:sstar-dyn}, would raise the predicted $(16,8)$ onset by $\ln(1/\numKapCfgRatio)/\ln(128) \approx +\numKapCfgPlateauShift$ against the observed residual of $+\numPlateauResidVOneSixDEight$ ($\numOnsetVOneSixDEight$ vs.\ $\numPredVOneSixDEight$), though within $(8,4)$ alone $\kappa$ varies by $\numKappaSSpread\times$ across the $\numKappaSCells$ stalled cells $s \in [4.5, 7.5]$. To test directly whether $\varepsilon_{\mathrm{dyn}}$ tracks per-cell $\kappa\alpha$ when all configurations are probed at matched distance above their walls ($s$ rounded to the nearest $0.5$ grid point to $\sstar(10^{-12}) + 0.5$: $s = 6.0, 5.0, 4.5, 4.0, 4.0, 7.0$; Column 8 of Table~\ref{tab:scales}), we measure the 5-seed geometric mean $\kappa$ in all six cells (Table~\ref{tab:kappa-test}). If $\varepsilon_{\mathrm{dyn}}$ scaled with per-cell $\kappa\alpha$, the plateau onset would shift by $\Delta_{\mathrm{pred}} = \ln(\kappa_{(8,4)}/\kappa_{(V,D)})/\ln(VD)$; Table~\ref{tab:kappa-test} sets this against the measured residual $\Delta_{\mathrm{meas}}$ from the constant-$\varepsilon_{\mathrm{dyn}}$ prediction of Table~\ref{tab:scales}. The decision rule was fixed before the run (Appendix~\ref{app:ledger}): on the $\numKapDiscCells$ non-reference $D \ge 4$ cells with $|\Delta_{\mathrm{pred}}| \ge 0.30$, where the two hypotheses are at least $0.30$ apart, a cell supports the constant scalar if $|\Delta_{\mathrm{meas}}| \le 0.15$ and supports per-cell $\kappa\alpha$ if $|\Delta_{\mathrm{pred}} - \Delta_{\mathrm{meas}}| \le 0.15$. All $\numKapVotesScalar$ of the $\numKapDiscCells$ cells support the constant scalar and $\numKapVotesTracks$ support per-cell $\kappa\alpha$ ($\numKapLiteralPassCells$ of $5$ when all five $D \ge 4$ cells, including the reference cell $(8,4)$, are counted). This establishes that the post-stall steady-state denominator $\kappa$ measured at $s \approx \sstar + 0.5$ is not the threshold-crossing floor $\varepsilon_{\mathrm{dyn}}$ that sets $\sstar$ in~\eqref{eq:sstar-dyn}, and deriving the exact pre-stall relation between these two observables remains open theoretical work (\S\ref{sec:discussion}, item 2).

\begin{table}[t]
\centering
\caption{Matched-offset test of whether $\varepsilon_{\mathrm{dyn}}$ tracks per-cell $\kappa\alpha$ ($\varepsilon_0 = 10^{-12}$, $n=5$ seeds). $\kappa$ is the 5-seed geometric mean of $\sqrt{\hat{v}_V}/\alpha$ at the probed $s$ (the grid point nearest $\sstar(10^{-12}) + 0.5$); $\Delta_{\mathrm{pred}}$ is the onset shift that per-cell $\kappa\alpha$ would cause, and $\Delta_{\mathrm{meas}}$ is the measured residual from the constant-$\varepsilon_{\mathrm{dyn}}$ prediction in Table~\ref{tab:scales}.}
\label{tab:kappa-test}
\small
\begin{tabular}{lrrrrr}
\toprule
$(V,D)$ & probed $s$ & $\kappa\ (10^{-6})$ & ratio to $(8,4)$ & $\Delta_{\mathrm{pred}}$ & $\Delta_{\mathrm{meas}}$ \\
\midrule
$(8,4)$  & 6.0 & \numKapCellVEightDFour     & \numKapRatioVEightDFour     & \numKapDeltaPredVEightDFour     & \numKapDeltaMeasVEightDFour \\
$(16,4)$ & 5.0 & \numKapCellVOneSixDFour    & \numKapRatioVOneSixDFour    & \numKapDeltaPredVOneSixDFour    & \numKapDeltaMeasVOneSixDFour \\
$(16,8)$ & 4.5 & \numKapCellVOneSixDEight   & \numKapRatioVOneSixDEight   & \numKapDeltaPredVOneSixDEight   & \numKapDeltaMeasVOneSixDEight \\
$(32,4)$ & 4.0 & \numKapCellVThreeTwoDFour  & \numKapRatioVThreeTwoDFour  & \numKapDeltaPredVThreeTwoDFour  & \numKapDeltaMeasVThreeTwoDFour \\
$(32,8)$ & 4.0 & \numKapCellVThreeTwoDEight & \numKapRatioVThreeTwoDEight & \numKapDeltaPredVThreeTwoDEight & \numKapDeltaMeasVThreeTwoDEight \\
\midrule
$(16,2)$ & 7.0 & \numKapCellVOneSixDTwo     & \numKapRatioVOneSixDTwo     & \numKapDeltaPredVOneSixDTwo     & \numKapDeltaMeasVOneSixDTwo \\
\bottomrule
\end{tabular}
\end{table}

\subsection{Six mechanistic interventions on the $\kappa\alpha$ floor}
\label{sec:interventions}

We now test six direct predictions of the cross-mode leakage mechanism~\eqref{eq:leakage}--\eqref{eq:vhat-floor}.

\textbf{1. Decoupled diagonal and Zipf-softmax controls: removing mode coupling restores the $\varepsilon_0$ lever.}
If cross-mode leakage through the off-diagonal terms of $A_k$ is what pins $\sqrt{\hat{v}_V}$, then replacing the coupled sensing loss with a decoupled coordinate objective should eliminate the floor and restore the $\varepsilon_0$ slope across $\varepsilon_0 \in \{10^{-8}, 10^{-10}, 10^{-12}\}$. Testing a Zipf-prior diagonal/softmax head at $V = 128$ ($n=5$ seeds) across $\varepsilon_0 \in \{10^{-8}, 10^{-10}, 10^{-12}\}$ confirms that without the Burer--Monteiro off-diagonal sensing coupling, the onset moves linearly with $\log_{10}(1/\varepsilon_0)$ right down to $10^{-12}$ with measured slope $\mathrm{d}\sstar/\mathrm{d}\log_{10}(1/\varepsilon_0) = \numSoftmaxSlopeMeasured$ against the parameter-free prediction $\ln(10)/\ln(128) = \numSoftmaxSlopePredicted$ (relative error $\numSoftmaxSlopeErrPct\%$).

\textbf{2. Varying $\beta_1$ and testing RMSProp ($\beta_1 = 0$).}
In~\eqref{eq:au} and~\eqref{eq:leakage}, the fast-mode flip amplitude scales as $A^\ast \propto (1-\beta_1)/(1+\beta_1)$ (spanning a factor of $\numBetaOneSpan\times$ across the $\numBetaOneArms$ momentum arms $\beta_1 \in \{0.5, 0.8, 0.9, 0.95, 0.99\}$). Sweeping $\beta_1 \in \{0, 0.5, 0.8, 0.9, 0.95, 0.99\}$ at $(8,4)$, $\varepsilon_0 = 10^{-12}$ ($n=5$ seeds) tests how the rare-row floor $r(\beta_1) := \sqrt{\hat{v}_V}/\alpha$ responds relative to $\beta_1 = 0.9$:
\begin{center}\small
\begin{tabular}{lrrrrrr}
\toprule
$\beta_1$ & $0$ (RMSProp) & $0.5$ & $0.8$ & $0.9$ (ref.) & $0.95$ & $0.99$ \\
\midrule
Predicted ratio $\frac{1-\beta_1}{1+\beta_1} / \frac{0.1}{1.9}$ & $\numRmspropPred$ & $\numBetaOneRatioHalfPred$ & $\numBetaOneRatioEightPred$ & $1.00$ & $\numBetaOneRatioNineFivePred$ & $\numBetaOneRatioNineNinePred$ \\
Measured $\sqrt{\hat{v}_V}/\alpha$ ratio (median stalled) & $\numRmspropRatio$ & $\numBetaOneRatioHalf$ & $\numBetaOneRatioEight$ & $1.00$ & $\numBetaOneRatioNineFive$ & $\numBetaOneRatioNineNine$ \\
Measured $60\text{k}$ onset $\sstar(10^{-12})$ & $\numBetaOneWallZero$ & $\numBetaOneWallHalf$ & $\numBetaOneWallEight$ & $\numBetaOneWallNine$ & $\numBetaOneWallNineFive$ & $\numBetaOneWallNineNine$ \\
\bottomrule
\end{tabular}
\end{center}
The two low-$\beta_1$ onsets come from an additional 50-run sweep over $s \in \{2.5, 3.0, 3.5, 4.0, 4.5\}$ with the same code as the original $\beta_1 = 0$ and $\beta_1 = 0.5$ runs, which start at $s = 4.5$, where both arms are already above threshold (Appendix~\ref{app:ledger}). Three features stand out. First, across the entire sweep $\beta_1 \in \{0, 0.5, 0.8, 0.9, 0.95, 0.99\}$, the measured floor itself decreases monotonically with $\beta_1$ from $\numRmspropRatio\times$ at $\beta_1 = 0$ down to $\numBetaOneRatioNineNine\times$ at $\beta_1 = 0.99$ (with $5/5$ seeds in the predicted direction on all $\numBetaOneDirArms$ arms stalled at $s=6.0$, and across all $\numRmspropPairs$ seed-paired comparisons at $\beta_1 = 0$), and the $60\text{k}$ onset $\sstar(10^{-12})$ increases monotonically from $\numBetaOneWallZero$ at $\beta_1 = 0$ and $\numBetaOneWallHalf$ at $\beta_1 = 0.5$ through $\numBetaOneWallEight$ at $\beta_1 = 0.8$, $\numBetaOneWallNine$ and $\numBetaOneWallNineFive$ to $\numBetaOneWallNineNine$ at $\beta_1 = 0.99$. Notice that RMSProp ($\beta_1 = 0$) exhibits a floor $\numRmspropRatio\times$ higher than Adam at $\beta_1 = 0.9$, confirming that RMSProp shares the self-generated floor even though its 1-D flip orbit has $|\mu|_{\max} = \numFloqZeroMuMid = \beta_2 < 1$. Second, the floor response is sub-linearly compressed relative to the 1-D factor $(1-\beta_1)/(1+\beta_1)$ (measured/predicted ratio $\numBetaOneCompZero$ at $\beta_1=0$ and $\numBetaOneCompLo\text{--}\numBetaOneCompHi$ across the four $\beta_1 > 0$ arms; through-origin log-log slope $\numBetaOneSlope$; the fit without the $\beta_1 = 0.5$ arm is in Table~\ref{tab:dispersion}). Physically, the exact $(1-\beta_1)/(1+\beta_1)$ prefactor is derived for a single isolated quadratic mode oscillating in a locked period-2 square wave; in the coupled $V \times D$ system, multi-mode interaction and phase jitter across burst-and-relaxation cycles attenuate the net momentum cancellation, causing the empirical floor response to deviate from the uncoupled 1-D square-wave scaling. Third, what reverses above $\beta_1 = 0.9$ is the measured-to-predicted compression ratio $[r(\beta_1)/r(0.9)]/[\frac{1-\beta_1}{1+\beta_1}/\frac{0.1}{1.9}]$---which climbs from $\numBetaOneCompZero$ at $\beta_1=0$ and $\numBetaOneCompLo$ at $\beta_1=0.5$ to $1.00$ at $\beta_1=0.9$, then turns back down to $\numBetaOneCompNineNine$ at $\beta_1 = 0.99$ (measured floor ratio $\numBetaOneRatioNineNine$ vs.\ predicted $\numBetaOneRatioNineNinePred$)---whereas both the floor itself and $\sstar$ remain strictly monotone through $\beta_1 = 0.99$. Similarly, varying $\beta_2 \in \{0.99, 0.999, 0.9999\}$ at $T = 500\text{k}$ shifts the duty-cycle wall only slightly across $[\numDutyWallHi, \numDutyWallLo]$ ($\numDutyWall$ at $\beta_2 = 0.999$), a span of $\numDutySpanApprox$ ($\numDutySpan$, within $\numDutyMargin$ of the $0.25$ band; Table~\ref{tab:dispersion}).

\textbf{3. Default learning rate $\alpha = 10^{-3}$.}
Equation~\eqref{eq:vhat-floor} predicts that lowering the step size $\alpha$ by $10\times$ from $10^{-2}$ to Adam's PyTorch default $\alpha = 10^{-3}$ should lower $\sqrt{\hat{v}_V}$ by $\sim 10\times$. At $(8,4)$, $s = 6.0$, $\varepsilon_0 = 10^{-12}$ ($n=5$ seeds), the paired ratio $\sqrt{\hat{v}_V}(\alpha{=}10^{-2})/\sqrt{\hat{v}_V}(\alpha{=}10^{-3})$ has seed median $\numDefaultLrPairRatio\times$. Relative to the default hyperparameter $\varepsilon_0 = 10^{-8}$, the $\alpha = 10^{-3}$ chatter floor $\rho := \sqrt{\hat{v}_V}(\alpha{=}10^{-3}) / 10^{-8}$ has seed median $\numDefaultLrRhoMedian$, with only $\numDefaultLrAbove$ of $5$ seeds above $10^{-8}$ (per-seed spread in Table~\ref{tab:dispersion}), confirming that lowering $\alpha$ to $10^{-3}$ pushes the dynamical floor below default $\varepsilon_0 = 10^{-8}$ on the majority of seeds.

\textbf{4. Second objective family: weighted symmetric matrix factorization.}
To verify that the self-generated floor is not an artifact of random Gaussian trace sensing matrices $A_k$, we test direct weighted symmetric factorization $\min_U \|\Sig^{1/2}(UU^\top - M^\ast)\Sig^{1/2}\|_F^2$ ($n=5$ matched seeds). Sweeping $\varepsilon_0$ from $10^{-10}$ to $10^{-12}$ yields $60\text{k}$-step onsets $\numFamTwoOnsetTen$ and $\numFamTwoOnsetTwelve$---a shift of $\Delta\sstar(10^{-12} - 10^{-10}) = \numFamTwoDelta$, while the rare-row denominator sits at a median of $\numFamTwoRhoMedian\times \varepsilon_0$ at $\varepsilon_0 = 10^{-8}$ ($\kappa = \numFamTwoKappa\times 10^{-9}$; dispersion in Table~\ref{tab:dispersion}), confirming a self-generated floor in a second matrix family.

\textbf{5. Mini-batch sampling: deterministic chatter vs.\ stochastic gradient variance.}
In full-batch optimization ($B = m = 48$ at $(8,4)$), the floor $\sqrt{\hat{v}_V} \approx \kappa\alpha$ comes entirely from deterministic cross-mode chatter near $M^\ast$. Under mini-batch sampling of batch size $B < m$ without replacement ($n=5$ matched seeds), a non-vanishing residual $\|UU^\top - M^\ast\|_F > 0$ also generates classical mini-batch sampling variance $\sigma_{\mathrm{mb}}(B) \propto \sqrt{(m-B)/(B(m-1))}$. Testing $B = 12$ and $B = 24$ at $(8,4)$ shows that the wall moves lower ($\sstar(10^{-12}) = \numMbWallTwelve$ at $B=12$ and $\numMbWallTwentyFour$ at $B=24$) while $\Delta\sstar(10^{-12} - 10^{-10})$ remains near zero ($\numMbDeltaTwelve$ at $B=12$ and $\numMbDeltaTwentyFour$ at $B=24$; Table~\ref{tab:dispersion}). The paired denominator ratio $\sqrt{\hat{v}_V}(B)/\sqrt{\hat{v}_V}(\text{full batch})$ is $\numMbRhoTwelve\times$ at $B=12$ ($\numMbRhoBothTwelve\times$ on shared stalled cells) and $\numMbRhoTwentyFour\times$ at $B=24$ ($\numMbRhoBothTwentyFour\times$ on shared stalled cells), because mini-batch gradient variance $\sigma_{\mathrm{mb}} \gg \kappa\alpha$ adds a larger $\varepsilon_0$-independent stochastic floor on top of deterministic chatter.

\textbf{6. Smaller step size $\alpha_1$ lowers the floor directly, while late step-size decay vs.\ constant $\alpha_1$ is non-monotone across $s$.}
Because $\varepsilon_{\mathrm{dyn}} \propto \alpha$, the active lever for lowering the self-generated floor is a smaller step size $\alpha_1$; reducing $\alpha$ mid-run quenches fast-mode period-2 oscillation and collapses $\sqrt{\hat{v}_V}$ relative to continuing at $\alpha = 10^{-2}$, while its performance relative to running at constant $\alpha_1$ from initialization depends on the exponent $s$.

\begin{figure}[t]
\centering
\includegraphics[width=0.46\linewidth]{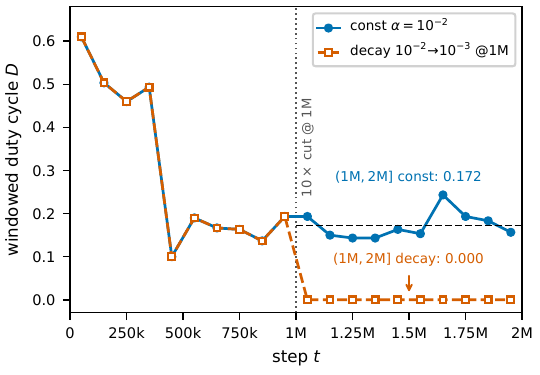}\hfill
\includegraphics[width=0.52\linewidth]{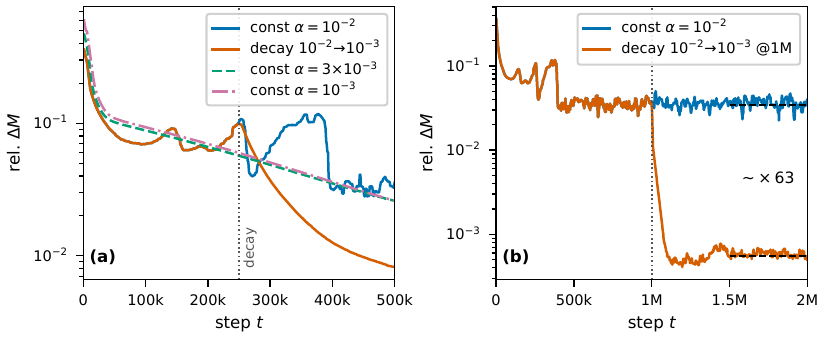}
\caption{\emph{Left:} Fast-mode sign-alternation duty cycle $D(t) = \Pr(\relDM > 0.05)$ in $100\text{k}$-step windows across $t \in [0, 2\times 10^6]$ before and after a $10\times$ learning-rate drop at step $10^6$ ($(V,D)=(8,4)$, $s=6.0$, $\varepsilon_0=10^{-12}$, $n=3$ seeds); reducing $\alpha$ to $10^{-3}$ quenches the $(1\text{M}, 2\text{M}]$ chatter duty cycle to $\numDutyPostDecay$ (vs.\ $\numDutyPostConst$ on constant $\alpha = 10^{-2}$). \emph{Right:} Recovery error $\relDM(t)$ for constant $\alpha = 10^{-2}$ vs.\ $10\times$ step-size decay ($\alpha: 10^{-2} \to 10^{-3}$) up to $2\times 10^6$ steps across three seeds; equal-budget constant-$\alpha_1$ controls show a non-monotone trade-off across the two measured exponents ($\numAlignRatioSix\times$ advantage for late decay at $s=6.0$, vs.\ $\numAlignWorseFive\times$ disadvantage at $s=5.5$).}
\label{fig:decay}
\end{figure}

Figure~\ref{fig:decay} tests this intervention at $(V,D) = (8,4)$, $\varepsilon_0 = 10^{-12}$, $s \in \{5.5, 6.0\}$ over $2\times 10^6$ steps, comparing constant $\alpha = 10^{-2}$ against step-size cuts to $\alpha_1 \in \{3\times 10^{-3}, 10^{-3}, 3\times 10^{-4}\}$ and against an equal-budget constant-$\alpha_1$ control:
\begin{itemize}
    \item Immediately after the $10\times$ step-size cut at step $10^6$, the fast-mode period-2 duty cycle in $100\text{k}$-step windows drops to $\numDutyPostDecay$ over $(1\text{M}, 2\text{M}]$ (peak post-cut $\relDM = \numDutyPostDecayPeak$, $\numDutyPostDecayMargin\%$ below the $0.05$ duty threshold, compared to $\numDutyPostConst$ on the constant-$\alpha$ arm).
    \item Compared against continuing at the constant $\alpha = 10^{-2}$ baseline to $2\times 10^6$ steps, the $10\times$ step-size cut reduces the settled $(1.5\text{M}, 2\text{M}]$ floor by $\numFloorRatio\times$ (from $\numFloorConst\times 10^{-2}$ down to $\numFloorDecayOneM\times 10^{-4}$, or $\numFloorOneE\times 10^{-4}$, $\numFloorThreeE\times 10^{-3}$, and $\numFloorThreeF\times 10^{-5}$ across the three decayed arms $\alpha_1 \in \{10^{-3}, 3\times 10^{-3}, 3\times 10^{-4}\}$, with per-seed constant-vs.-decayed final ratios spanning $\numDecaySepLo\text{--}\numDecaySepHi\times$ and seed-paired ratios $\numDecayPairedSixA\times, \numDecayPairedSixB\times, \numDecayPairedSixC\times$ at $s=6.0$).
    \item To separate the benefit of a smaller constant step size $\alpha_1$ from the schedule transition itself, we also compare against an equal-budget ($T = 2\times 10^6$) constant-$\alpha_1$ null control ($n=5$ seeds, closing $\numAlignGapFive\%$ and $\numAlignGapSix\%$ of the Adam--Newton gap at $s=5.5$ and $6.0$). Across the two measured exponents, the relative performance of late step-size decay vs.\ equal-budget constant $\alpha_1$ is non-monotone: at $s = 6.0$, the late-drop schedule outperforms the equal-budget constant-$\alpha_1$ null by a ratio of medians of $\numAlignRatioSix\times$ ($\numAlignDexSix$ dex; paired geometric mean $\numAlignPairedSix\times$ with $\numAlignPairedSixDir/5$ seeds favouring decay), whereas at $s = 5.5$, the equal-budget constant-$\alpha_1$ null reaches a lower final residual than the late-drop schedule (ratio of medians $\numAlignRatioFive\times$, i.e.\ the late drop is $\numAlignWorseFive\times$ higher than running at $\alpha_1$ from the start, $+\numAlignDexFive$ dex; paired geometric mean $\numAlignPairedFive\times$ with $\numAlignPairedFiveDir/5$ seeds $> 1$). Thus the relative advantage of late decay over constant $\alpha_1$ reverses between the two tested exponents $s = 5.5$ and $s = 6.0$ depending on how far the instance sits above the wall (Appendix~\ref{app:ledger}, item 3 gives the full breakdown).
\end{itemize}

\section{Discussion and Limitations}
\label{sec:discussion}

\textbf{Why $\varepsilon_0 = 10^{-8}$ is a special operating point.}
A striking practical consequence of~\eqref{eq:vhat-floor} is that at Adam's standard learning rate range $\alpha \in [10^{-3}, 10^{-2}]$ and $\beta_1 = 0.9$, the self-generated floor $\varepsilon_{\mathrm{dyn}} \sim 5\times 10^{-10}\text{--}5\times 10^{-9}$ sits just below the default hyperparameter $\varepsilon_0 = 10^{-8}$. Consequently, experiments that increase $\varepsilon_0$ above $10^{-8}$ (e.g.\ $10^{-6}$ or $10^{-4}$) observe clean hyperparameter sensitivity, whereas experiments that lower $\varepsilon_0$ below $10^{-8}$ hit the dynamical chatter floor and see no improvement. In realistic stochastic training, mini-batch sampling variance $\sigma_{\mathrm{mb}}$ typically exceeds the deterministic $\kappa\alpha$ chatter floor once batch noise is present ($\numMbRhoTwelve\times$ and $\numMbRhoTwentyFour\times$ larger rare-row denominator at $B = 12$ and $B = 24$ in \S\ref{sec:interventions}, item 5), providing an even higher $\varepsilon_0$-independent floor. Thus deterministic $\kappa\alpha$ chatter is the dominant floor in full-batch $\mathrm{fp64}$ optimization on small-to-medium testbeds, whereas under mini-batch sampling or low-precision ($\mathrm{fp16}/\mathrm{bf16}$) quantization, stochastic gradient noise or mantissa quantization dominates the effective floor. This reconciles two seemingly contradictory observations in the literature: why tuning $\varepsilon_0$ matters in large-scale training when $\varepsilon_0$ is too large or gradients underflow in low precision~\citep{everett2024adam}, and why shrinking $\varepsilon_0$ toward zero in high precision fails to rescue ill-conditioned tail modes unless accompanied by a smaller learning rate or higher-order preconditioning~\citep{cohen2022adaptive,kunstner2024zipf}.

\textbf{Limitations and scope boundaries.}
\begin{enumerate}
    \item \textbf{Controlled matrix sensing and factorization substrate.} All experiments in this paper are conducted on rank-$D$ Burer--Monteiro matrix sensing and weighted factorization ($V \le 64$) where the exact expected Hessian spectrum and ground-truth recovery error $\relDM$ can be computed analytically and solved to machine precision by Gauss--Newton. In deep neural networks, layerwise normalization, stochastic data augmentation, and time-varying feature kernels introduce additional gradient variance sources beyond the deterministic cross-mode coupling coefficient $\kappa$. Measuring the relative magnitudes of fast-mode edge-of-stability chatter and mini-batch gradient noise in transformer language models is an important direction for future work.
    \item \textbf{Phenomenological coupling constant $\kappa$, multi-mode persistence, scale transfer, and sampling variability.} While the 1-D period-2 orbit amplitude $A(u) = \frac{\alpha}{38}(1 - 38/u)$ is exact (Proposition~\ref{thm:flip}), the persistence of fast-mode oscillations in the coupled $VD$-dimensional system is established here via numerical duty-cycle measurements and phase-incommensurability arguments rather than a formal multi-mode Floquet theorem, and the coupling constant $\kappa$ in $\sqrt{\hat{v}_V} \approx \kappa\alpha$ is measured empirically rather than derived in closed form; formalizing multi-mode Floquet persistence is left to future work. Across the six $(V,D)$ configurations in Table~\ref{tab:scales}, probing $\sqrt{\hat{v}_V}$ at matched distance $s \approx \sstar(10^{-12}) + 0.5$ above each wall (we probe only the single offset $+0.5$ above $\sstar$: sensitivity of $\kappa$ to the offset choice is not characterized, and in particular $\sqrt{\hat{v}_V}/\alpha$ is not measured at or below $\sstar$, where runs still converge) shows that while the post-stall rare-row denominator $\kappa$ decreases with $(V,D)$ from $\numKapCellVEightDFour\times 10^{-6}$ at $(8,4)$ down to $\numKapCellVThreeTwoDEight\times 10^{-6}$ at $(32,8)$, the $D \ge 4$ threshold crossing $\sstar(10^{-12})$ is governed by a wall-equivalent single scalar $\varepsilon_{\mathrm{dyn}}$ ($\numKapVotesScalar$ of $\numKapDiscCells$ discriminating cells; \S\ref{sec:scale-transfer}); we do not provide a theoretical explanation for why threshold crossing integrates out this cell-by-cell variation in post-stall $\kappa$, and deriving the exact pre-stall relation between the threshold-crossing floor $\varepsilon_{\mathrm{dyn}}$ and the post-stall chaotic steady state $\kappa\alpha$ is an important direction for future theoretical work. Across the $20$-seed budget ensemble at $600\text{k}$ steps, the 5-of-20 subset onset range spans $\numWorstInstability$ exponent units (making the headline gaps $\numGapOverInstabilityLo\text{--}\numGapOverInstabilityHi\times$ the worst 600k subset instability, though $> 15\times$ the matched-seed $\varepsilon_0$ difference).
    \item \textbf{Finite-$D$ spectral crossover and $D = 2$ breakdown.} The underlying curvature scaling $\lambda_{\min}^{\mathrm{ng}} \propto V^{-(s+1)}D^{-s}$ is a large-$D$ leading-order approximation (Appendix~\ref{app:companion-wall}) that holds for $D \ge 4$ and breaks down at $D = 2$ (Table~\ref{tab:scales}); we do not derive an analytical bound for the finite-$D$ transition between Law A and Law B, and testing whether the companion paper's finite-$D$ Law A spectral correction quantitatively explains the $+\numResidDtwo$ residual at $(16,2)$ is left for follow-up work.
\end{enumerate}

\section{Conclusion}
\label{sec:conclusion}

Lowering Adam's denominator constant $\varepsilon_0$ below $10^{-8}$ in 64-bit arithmetic fails to shift the power-law precision wall not because of arithmetic roundoff, spurious local minima, or finite training budgets, but because Adam generates its own internal denominator floor $\varepsilon_{\mathrm{dyn}} \propto \alpha$. Fast modes oscillating near the adaptive edge of stability ($u_{c2} = 2(1+\beta_1)/(1-\beta_1) = 38$) leak gradient fluctuations of scale $\kappa\alpha$ into nearly flat rare modes through measurement coupling, pinning $\sqrt{\hat{v}_V} \approx \kappa\alpha \gg \varepsilon_0$. Calibrating a single threshold-crossing wall-equivalent floor parameter $\varepsilon_{\mathrm{dyn}} = \numEpsDyn \times 10^{-9}$ at $\alpha = 10^{-2}$ predicts the saturated failure boundary across $D \ge 4$ problem scales with RMS $\numRmsDge$, and six targeted interventions test the factors of the mechanism (with the $\beta_1$ response matching the predicted monotone direction while exhibiting sub-linear amplitude compression, $\numBetaOneCompZero$ at $\beta_1=0$ and $\numBetaOneCompLo\text{--}\numBetaOneCompHi$ for $\beta_1 > 0$). While shrinking $\varepsilon_0$ is ineffective below $10^{-8}$, operating at a smaller step size $\alpha_1$ lowers the chatter floor directly: mid-run $10\times$ step-size decay yields a $\numDecaySepLo\text{--}\numDecaySepHi\times$ lower stalled residual than continuing at $\alpha = 10^{-2}$, whereas relative to an equal-budget constant-$\alpha_1$ control run from initialization, late decay is non-monotone across the two tested exponents---improving by $\numAlignRatioSix\times$ at $s = 6.0$ but trailing by $\numAlignWorseFive\times$ ($\numAlignRatioFive\times$) at $s = 5.5$.

\bibliographystyle{plainnat}

\newpage
\appendix

\section{Mathematical Proofs and Companion Precision-Wall Derivation}
\label{app:proofs}

\subsection{Proof of Proposition~\ref{thm:flip} and Observation~\ref{obs:flip}}
\label{app:proof-flip}

\begin{proof}[Proof of Proposition~\ref{thm:flip}]
Consider the warm-started 1-D Adam recurrence on $f(x) = \frac{1}{2}\lambda x^2$ ($g_t = \lambda x_t$):
\[
m_t = \beta_1 m_{t-1} + (1-\beta_1)\lambda x_t, \qquad
v_t = \beta_2 v_{t-1} + (1-\beta_2)\lambda^2 x_t^2, \qquad
x_{t+1} = x_t - \alpha\frac{m_t}{\sqrt{v_t} + \varepsilon_0}.
\]
\emph{Part 1 (Period-2 orbit existence and amplitude).} Along a period-2 alternating orbit $x_t = (-1)^t A$ with $A > 0$, we have $x_t^2 = A^2$ for all $t$, so the unique stationary point of the $v_t$ recurrence is $v_t = \lambda^2 A^2$ ($\sqrt{v_t} = \lambda A$). Similarly, the unique period-2 solution of $m_t = \beta_1 m_{t-1} + (1-\beta_1)\lambda x_t$ satisfying $m_t = -m_{t-1}$ is
\[
m_t = \frac{1-\beta_1}{1+\beta_1}\,\lambda x_t.
\]
Substituting $m_t$ and $\sqrt{v_t} = \lambda A$ into the position update $x_{t+1} = -x_t$ gives
\[
-x_t = x_t - \alpha\,\frac{1-\beta_1}{1+\beta_1}\frac{\lambda x_t}{\lambda A + \varepsilon_0}
\quad\Longleftrightarrow\quad
2 = \frac{\alpha(1-\beta_1)}{1+\beta_1}\frac{\lambda}{\lambda A + \varepsilon_0} = \frac{2A^\ast \lambda}{\lambda A + \varepsilon_0},
\]
where $A^\ast = \frac{\alpha(1-\beta_1)}{2(1+\beta_1)} = \frac{\alpha}{u_{c2}}$. Solving for $A$ gives $A = A^\ast - \varepsilon_0/\lambda = A^\ast(1 - u_{c2}/u)$, which is strictly positive if and only if $u = \alpha\lambda/\varepsilon_0 > u_{c2}$, i.e., $\lambda > \lambda_{\mathrm{stall}} = u_{c2}\varepsilon_0/\alpha$.

\emph{Part 2 (Stability of the origin for $u < u_{c2}$).} Near $(x,m,v) = (0,0,0)$, $\sqrt{v_t} = O(|x_t|) \ll \varepsilon_0$, so the update linearises to the constant-denominator system $m_t = \beta_1 m_{t-1} + (1-\beta_1)\lambda x_t$, $x_{t+1} = x_t - (\alpha/\varepsilon_0)m_t = x_t - u(m_t/\lambda)$, whose characteristic polynomial is
\[
P(z) = z^2 - \bigl(1 + \beta_1 - (1-\beta_1)u\bigr)z + \beta_1 = 0.
\]
By the Jury stability criteria ($P(1) = (1-\beta_1)u > 0$, $P(-1) = 2(1+\beta_1) - (1-\beta_1)u > 0$, and $|P(0)| = \beta_1 < 1$), both roots lie strictly inside the unit circle for $0 < u < u_{c2} = \frac{2(1+\beta_1)}{1-\beta_1}$, with real positive roots (monotone decay) when the discriminant $\bigl(1+\beta_1 - (1-\beta_1)u\bigr)^2 - 4\beta_1$ is non-negative, i.e.\ $0 < u \le u_{c1} := \frac{(1-\sqrt{\beta_1})^2}{1-\beta_1}$ ($u_{c1} \approx 0.0263$ at $\beta_1 = 0.9$), and complex conjugate roots inside the unit circle (damped oscillations) for $u_{c1} < u < u_{c2}$. At $u = u_{c2}$, a root exits the unit circle through $z = -1$ via a period-doubling (flip) bifurcation.

\emph{Part 3 (Transverse Floquet multipliers for $u > u_{c2}$).} Let $s := \lambda A / (\lambda A + \varepsilon_0) = 1 - u_{c2}/u \in (0,1)$. Linearising the two-step map $F^2$ around the period-2 cycle $(x, m, v) = (\pm A, \pm \frac{1-\beta_1}{1+\beta_1}\lambda A, \lambda^2 A^2)$ decouples perturbations into even/odd symmetry blocks. At $\beta_2 \to 1$ ($1-\beta_2 = 10^{-3}$), the $v$-multiplier is $\mu_{\mathrm{rad}} = \beta_2^2 + O((1-\beta_2)s) < 1$, while the coupled $3\times 3$ Floquet matrix $DF^2$ has spectral radius $|\mu|_{\max} = \numMuMid > 1$ at $(\beta_1,\beta_2) = (0.9, 0.999)$ for $u \gg u_{c2}$ ($\numMuLo$ at $\beta_2=0.99$ and $\numMuHi$ at $\beta_2=0.9999$, crossing $1$ at $\beta_1 \in [\numFloqCrossLo, \numFloqCrossHi]$). When $\beta_1 = 0$ (RMSProp), the $m_t$ state vanishes, the 2-D $(x,v)$ flip orbit has $x_{t+1} = -x_t$ at $u > 2$, and the two-step Jacobian has eigenvalues $\beta_2^2$ and $\beta_2 = \numFloqZeroMuMid < 1$.
\end{proof}

\textbf{Details of Observation~\ref{obs:flip} (two-stage relaxation and slow chatter).}
When the transverse instability in Part 3 of Proposition~\ref{thm:flip} drives $(x_t, m_t)$ out of anti-phase on a decoupled 1-D mode, $|x_t|$ drops sharply while $v_t$ remains large. During this quiescent phase ($\lambda x_t^2 \ll v_t$), $v_t \approx \beta_2 v_{t-1}$ decays at rate $\beta_2$ per step so $\sqrt{v_t} \propto \beta_2^{t/2}$ decays at rate $\beta_2^{1/2}$, producing a two-stage relaxation trajectory before the effective step size $\alpha\lambda/(\sqrt{v_t}+\varepsilon_0)$ rises back above $u_{c2}$ to ignite the next flare.

\subsection{Self-Contained Companion Derivation: Expected Hessian and Precision Wall}
\label{app:companion-wall}

To make this paper completely self-contained without requiring the reader to consult~\citet{companion2026wall}, we provide here the derivation of the expected Burer--Monteiro Hessian, the finite-$D$ spectral crossover measurements, and the 4-cell calibration of $\delta_{\mathrm{Adam}} = 0.482$ used in~\eqref{eq:wall} and Table~\ref{tab:scales}.

\textbf{1. Exact expected Hessian at $U^\ast$.}
At any ground-truth factor $U^\ast \in \R^{V \times D}$ ($M^\ast = U^\ast {U^\ast}^\top$), all residuals $r_k(U^\ast) = 0$ vanish and the sample Hessian of $L(U) = \frac{1}{m}\sum_{k=1}^m (\langle A_k, UU^\top\rangle - b_k)^2$ acting on $\delta U \in \R^{V \times D}$ is
\[
H_s\bigl|_{U^\ast}(\delta U) = \frac{8}{m} \sum_{k=1}^m \langle A_k U^\ast, \delta U\rangle\, A_k U^\ast.
\]
For $A_k = \Sig^{1/2} G_{\mathrm{sym},k} \Sig^{1/2}$ with $G_{\mathrm{sym},k} = (G_k + G_k^\top)/2$ and $G_{k,ij} \stackrel{\mathrm{iid}}{\sim} \mathcal{N}(0, 1/(Vm))$, the second moment $\E[G_{\mathrm{sym},ip} G_{\mathrm{sym},jq}] = (\delta_{ij}\delta_{pq} + \delta_{iq}\delta_{pj})/(2Vm)$ yields
\begin{equation}
\label{eq:app-hessian}
\E\bigl[H_s \mid U^\ast\bigr] = \frac{4}{Vm}\bigl(\Sig \otimes N^\ast + \mathcal{T}\bigr), \qquad
N^\ast := {U^\ast}^\top \Sig U^\ast \in \R^{D \times D},
\end{equation}
where $\mathcal{T}(\delta U) = \Sig U^\ast (\delta U)^\top \Sig U^\ast$ has rank at most $D^2$. On the $D(D-1)/2$-dimensional orthogonal gauge tangent space $\delta U = U^\ast \Omega$ ($\Omega^\top = -\Omega$), $(\Sig \otimes N^\ast)(\delta U) = \Sig U^\ast \Omega N^\ast = -\mathcal{T}(\delta U)$, producing $D(D-1)/2$ exact zero eigenvalues.

\textbf{2. Leading-order non-gauge eigenvalue and the finite-$D$ spectral crossover.}
Transverse to the gauge modes, the bottom of the spectrum of $\E[H_s \mid U^\ast]$ is carried by the Kronecker factor $\frac{4}{Vm}\lambda_{\min}(\Sig)\nu_{\min}(N^\ast)$, where $\nu_{\min}(N^\ast) := \lambda_{\min}(N^\ast)$ and $\lambda_{\min}(\Sig) = V \cdot V^{-s}/\sum_{j=1}^V j^{-s} \sim V^{1-s}$. With $m = 1.5VD$, the prefactor scales as $\frac{4}{Vm} \propto V^{-2}D^{-1}$. For isotropically drawn $U^\ast \sim \mathcal{N}(0, I/V)$, $N^\ast = \sum_{j=1}^V \Sigma_{jj} u_j u_j^\top$ is a sum of $V$ rank-one matrices in $\R^{D \times D}$ with $\|u_j\|^2 \sim D/V$:
\begin{itemize}
    \item Under the large-$D$ tail-mass approximation $\nu_{\min}(N^\ast) \sim \sum_{j > D} \Sigma_{jj}\|u_j\|^2 \sim D^{1-s}$ ("Law B"), combining $\frac{4}{Vm}\lambda_{\min}(\Sig)\nu_{\min}(N^\ast)$ yields the leading-order scaling
    \[
    \lambda_{\min}^{\mathrm{ng}} \;\propto\; V^{-2}D^{-1} \cdot V^{1-s} \cdot D^{1-s} \;=\; V^{-(s+1)}D^{-s}.
    \]
    \item Under a small-$D$ single-weight approximation $\nu_{\min}(N^\ast) \sim D^{-s}$ ("Law A"), one obtains $\lambda_{\min}^{\mathrm{ng}} \propto (VD)^{-(s+1)}$.
\end{itemize}
Evaluating $\nu_{\min}(N^\ast)$ and the carrier $\frac{4}{Vm}\Sigma_{VV}\nu_{\min}(N^\ast)$ numerically across $120$ exact spectral configurations ($V \in \{8,\ldots,256\}$, $D \in \{2,\ldots,32\}$, $s \in \{2,\ldots,6\}$) and $44$ finite-$m$ direct Hessian configurations ($\lambda_{\min}/\lambda_{\max} > 10^{-12}$) shows that $\nu_{\min}(N^\ast)$ is independent of $V$ ($\ln V$ slope $+0.297 - 0.041\,s$), while its $\ln D$ slope across this finite grid is $+0.501 - 1.069\,s$ (giving carrier exponent $V^{-(s+1)}D^{-0.501 - 1.069 s} \approx V^{-(s+1)}D^{-(s+1/2)}$, and direct finite-$m$ $\ln D$ coefficient $-0.048 - 1.146\,s$, which crosses the midpoint $-0.5 - s$ at $s = 3.10$, or $s = 3.0$ from the two-decimal rounded form $-0.05 - 1.15\,s$). On a wider product grid up to $D = 64$, the local $\ln D$ intercept $b_0 = \mathrm{d}\ln\nu_{\min}/\mathrm{d}\ln D + s$ transitions monotonically from $b_0 \approx 0.20$ at $D = 2\to 4$ (near Law A, $b_0=0$) to $b_0 = 1.019 \pm 0.048$ at $D = 32\to 64$ (Law B, $b_0=1$).

\textbf{3. Companion precision-wall formula and 4-cell calibration of $\delta_{\mathrm{Adam}} = 0.482$.}
Equating the leading-order non-gauge curvature $V^{-(s+1)}D^{-s}$ to the effective denominator floor $\varepsilon_{\mathrm{eff}}$ gives $s\log(VD) = \log(1/\varepsilon_{\mathrm{eff}}) - \log V$, and adding the metric-dependent detection lag $\delta_{\mathrm{Adam}}$ yields~\eqref{eq:wall}:
\[
\sstar(V, D) = \frac{\log(1/\varepsilon_{\mathrm{eff}}) - \log V}{\log(VD)} + \delta_{\mathrm{Adam}}.
\]
In~\citet{companion2026wall}, $\delta_{\mathrm{Adam}}$ is calibrated at $\varepsilon_0 = 10^{-8}$ ($\log(1/\varepsilon_0) = 18.421$) on four $D \ge 4$ configurations ($n=10$ seeds, $60\text{k}$ steps, rarest-direction onset threshold $0.05$):
\begin{center}\small
\begin{tabular}{rr|r|rrr}
\toprule
$V$ & $D$ & $m = 1.5VD$ & Base $\frac{\log(10^8) - \log V}{\log(VD)}$ & Measured onset ($\varepsilon_0 = 10^{-8}$) & Implied lag $\delta_i$ \\
\midrule
8  & 4 & 48  & $4.715$ & $5.208$ & $+0.493$ \\
16 & 8 & 192 & $3.225$ & $3.702$ & $+0.477$ \\
32 & 8 & 384 & $2.697$ & $3.194$ & $+0.497$ \\
16 & 4 & 96  & $3.763$ & $4.226$ & $+0.463$ \\
\midrule
\multicolumn{3}{l}{Mean calibrated lag $\delta_{\mathrm{Adam}}$} & \multicolumn{3}{l}{$\mathbf{0.482}$ (in-sample RMS $0.013$, max $|\Delta| \le 0.02$)} \\
\bottomrule
\end{tabular}
\end{center}
This calibrated constant $\delta_{\mathrm{Adam}} = 0.482$ is frozen throughout \S\ref{sec:validation} and Table~\ref{tab:scales}. It is calibrated on a different observable from the one fitted in this paper: the companion's rarest-direction onset at $\varepsilon_0 = 10^{-8}$ ($n = 10$; $5.208$ at $(8,4)$ in the table above), whereas this paper's $\relDM$ onset at the same $(8,4)$ and $\varepsilon_0$ is $\numOnsetEpsEight$ ($n = 5$). Section~\ref{sec:validation} reports how the fitted $\varepsilon_{\mathrm{dyn}}$ changes when the lag is calibrated on this paper's own $\varepsilon_0$-dominated rungs instead ($\delta = \numDeltaBranch$).

\subsection{Heuristic directional scaling of the leakage coefficient $\kappa$ (heuristic)}
\label{app:kappa-heuristic}

To provide an intuitive semi-analytical explanation for why the normalized rare-row denominator $\kappa := \sqrt{\hat{v}_V}/\alpha$ varies across problem dimensions $(V,D)$ and spectral exponent $s$ rather than remaining a universal constant, we trace the dimensional factors in~\eqref{eq:coupled-grad}--\eqref{eq:leakage}. We emphasize that the following argument is \textbf{heuristic} (not a formal theorem), because the residuals $r_k(U)$ are coupled to the sensing matrices $A_k$ and the number of active fast modes above $\lambda_{\mathrm{stall}}$ changes with $(V,D,s)$.

In~\eqref{eq:coupled-grad}, the gradient on the rarest row $i = V$ is
\[
g_V(U) = \frac{4}{m}\sum_{k=1}^m r_k(U)\,\sum_{j=1}^V (A_k)_{Vj} U_{j,:}, \qquad A_k = \Sig^{1/2} G_{\mathrm{sym},k} \Sig^{1/2}.
\]
When the common rows ($j \ll V$, $\lambda_j \gg \lambda_{\mathrm{stall}}$) oscillate near their period-2 / quasi-periodic attractor with displacement $\|\delta U_{\mathrm{fast}}\| \propto A^\ast = \frac{\alpha(1-\beta_1)}{2(1+\beta_1)}$, they drive shared residuals $r_k(U)$ of scale $O\!\bigl(\alpha\frac{1-\beta_1}{1+\beta_1}\bigr)$. Each off-diagonal entry $(A_k)_{Vj} = \Sigma_{VV}^{1/2}\Sigma_{jj}^{1/2} G_{\mathrm{sym},k,Vj}$ coupling a fast row $j$ into the rarest row $V$ carries two scale factors:
\begin{enumerate}
    \item the rare-row spectral weight $\Sigma_{VV}^{1/2} = \bigl(V \cdot V^{-s} / \sum_{\ell=1}^V \ell^{-s}\bigr)^{1/2} \sim \zeta(s)^{-1/2} V^{(1-s)/2}$, and
    \item the Gaussian entry scale $\mathrm{std}(G_{\mathrm{sym},k,Vj}) \propto (Vm)^{-1/2}$.
\end{enumerate}
Treating the $m = 1.5VD$ sensing projections $\frac{1}{m}\sum_{k=1}^m r_k(U)(A_k U)_{V,:}$ as an average over $m$ weakly correlated contributions across the $D$ coordinates of row $V$ yields the \textbf{heuristic} order-of-magnitude scaling
\begin{equation}
\label{eq:kappa-heuristic}
\kappa(V, D, s) \;:=\; \frac{\sqrt{\hat{v}_V}}{\alpha} \;\sim\; \frac{\|g_V\|_{\mathrm{leak}}}{\alpha}
\;\propto\; \frac{1-\beta_1}{1+\beta_1}\,\Sigma_{VV}^{1/2}\,V^{-1}\,D^{-1}
\;\propto\; \frac{1-\beta_1}{1+\beta_1}\,V^{-(s+1)/2}\,D^{-1},
\end{equation}
up to a slowly varying factor from the count and phase coherence of fast modes with $\lambda_j > \lambda_{\mathrm{stall}}$.

Despite its heuristic nature,~\eqref{eq:kappa-heuristic} accounts for the direction of every empirical dependence of $\kappa$ reported in \S\ref{sec:mechanism} and \S\ref{sec:interventions}:
\begin{itemize}
    \item \emph{Decrease with $s$ at fixed $(V,D)$:} Through $\Sigma_{VV}^{1/2} \propto V^{-s/2}$ (together with the shrinking pool of fast modes above $\lambda_{\mathrm{stall}}$ as $s$ grows), $\kappa$ decreases with $s$, matching the $\numKappaSSpread\times$ span observed across the $\numKappaSCells$ stalled exponent cells $s \in [4.5, 7.5]$ at $(8,4)$.
    \item \emph{Decrease with $(V,D)$:} Increasing $(V,D)$ from $(8,4)$ to $(16,8)$ reduces both the rare-row spectral factor $V^{(1-s)/2}$ and the per-measurement coupling factor $(Vm)^{-1/2}$, consistent with the measured drop in $\kappa$ from $\numKappaCurrentSix \times 10^{-6}$ at $(8,4)$ ($s=6.0$, re-run; $\numKappa \times 10^{-6}$ in the original run) to $\numKapCfgSixteenEight \times 10^{-6}$ at $(16,8)$ ($s \in \{4.5, 5.0\}$, cross-configuration ratio $\numKapCfgRatio$).
\end{itemize}

\section{Pre-registration Ledger and Protocol History}
\label{app:ledger}
\sloppy

This appendix documents the protocol history, the adjudication of the five initial hypotheses (H1--H5) and pre-registered experimental protocols (P1--P45), the reconciliation between stored-artifact and current-code evaluation paths on the six multi-scale configurations, the leave-one-seed-out vs.\ 20-seed bootstrap comparison, the four summary statistics for the step-size decay intervention, and the provenance of the macros in \texttt{numbers.tex}. \S\ref{app:sources} maps each main-text result to its protocol, and Table~\ref{tab:dispersion} collects the dispersion intervals and alternative fits that the main text omits.

\subsection{Adjudication of hypotheses H1--H5 and protocols P1--P45}

\begin{table}[h]
\centering\footnotesize
\setlength{\tabcolsep}{4pt}
\caption{Summary ledger of hypotheses H1--H5 and experimental protocols P1--P45.}
\label{tab:ledger-h}
\begin{tabular}{ll p{2.3cm} p{8.8cm}}
\toprule
ID & Hypothesis & Protocols & Outcome and Key Quantitative Finding \\
\midrule
H1 & Underflow / ULP & P1, P30, P42, P44 & \textbf{Refuted.} $\mathrm{fp64}$ sweep at $(8,4)$ gives onsets $\numOnsetEpsTwo$ ($10^{-2}$, P44), $\numOnsetEpsFour$ ($10^{-4}$), $\numOnsetEpsSix$ ($10^{-6}$), $\numOnsetEpsEight$ ($10^{-8}$), $\numOnsetEpsTen$ ($10^{-10}$), and $\numOnsetEpsTwelve$ ($10^{-12}$, de-censored shift $\numPOneDelta$; log-minus-linear max shift $\numLadderShiftMax$); at $(16,8)$, onsets are $\numOnsetVOneSixDEightFour$ ($10^{-4}$), $\numOnsetVOneSixDEightSix$ ($10^{-6}$, P42), $\numOnsetVOneSixDEightTen$ ($10^{-10}$), and $\numOnsetVOneSixDEight$ ($10^{-12}$); $10^{-10}\to 10^{-12}$ move is $\numSatMoveMeasured$ ($\numSatMoveMeasuredLog$ under log interpolation) at $(8,4)$ and $\numSatMoveMeasuredB$ at $(16,8)$. ULP roundoff SNR exceeds $10^8$. \\
H2 & Landscape traps & P2 & \textbf{Refuted.} Warm-started from Adam's $\numLbfgsN$ stalled $60\text{k}$ checkpoints ($\relDM \in [\numAdamResidLo, \numAdamResidHi]$), L-BFGS converges on $\numLbfgsConverged/\numLbfgsN$ ($\relDM \le \numLbfgsRelMax\times 10^{-10}$, $\numLbfgsStepsLo\text{k}\text{--}\numLbfgsStepsHi\text{k}$ steps) and damped Gauss--Newton on $12/\numLbfgsN$. \\
H3 & Finite step budget & P3, P27--P29, P34, P37 & \textbf{Refuted as an $\varepsilon_0$ explanation; power-law creep documented.} On the current path ($n=10$), $\Delta\sstar(10^{-12}-10^{-10})$ is $\numPathTenDeltaSixtyK$ ($60\text{k}$), $\numPathTenDeltaTwoHundredK$ ($200\text{k}$), and $\numPathTenDeltaSixHundredK$ ($600\text{k}$). Across $20$ seeds ($\numBootNSixtyK$, $\numBootNTwoHundredK$, $\numBootNSixHundredK$ 5-seed subsets, $\numBootCensoredSixtyK$, $\numBootCensoredTwoHundredK$, $\numBootCensoredSixHundredK$ censored subsets dropped), 5-seed subset SDs are $\numBootSdSixtyK$, $\numBootSdTwoHundredK$, $\numBootSdSixHundredK$ (95\% widths $\numBootRangeSixtyK$, $\numBootRangeTwoHundredK$, $\numBootRangeSixHundredK$). \\
H4 & 1-D flip orbit & P5, P9, P10, P12, P14, P15, P41 & \textbf{Entry point only; two 1-D limits disclosed.} Exact 1-D orbit $A(u) = \frac{\alpha}{38}(1-38/u)$ for $u > 38$, tracked by $A_{\mathrm{RMS}} \approx \numAuTrack A^\ast(1-38/u)$ within $\numAuErr\%$ (cross-leg collapse $\numAuConjSpread\%$ in P41). However, 1-D $\lambda_{\mathrm{stall}} = 38\varepsilon_0/\alpha \propto \varepsilon_0$ and $|\mu|_{\max} = \numMuMid > 1$, so 1-D alone cannot sustain an $\varepsilon_0$-independent floor. \\
H5 & Cross-mode leak & P4, P6, P7, P16, P17, P32--P40, P43, P45 & \textbf{Confirmed with domain boundaries.} At $(8,4)$, $\sqrt{\hat{v}_V}/\alpha = \numKappa\times 10^{-6}$ ($\sqrt{\hat{v}_V} = \numSelfEps\times 10^{-8}$). Fitted $\varepsilon_{\mathrm{dyn}} = \numEpsDyn\times 10^{-9}$ matches $D \ge 4$ onsets with RMS $\numRmsDge$ ($\numRmsDgeRerun$ on reruns); $D=2$ breaks down ($+\numResidDtwo$). Six-cell $\kappa$ probe (P43) confirms \texttt{\numKapTrackVerdict} ($\numKapVotesScalar/\numKapDiscCells$). Verified across P16, P32, P33, P35, P36, P39, P40, and P43. \\
\bottomrule
\end{tabular}
\end{table}

\subsection{Where the main-text results come from}\label{app:sources}

In the main text, ``original runs'' are the stored artifacts and ``re-runs'' are runs of the current released code (\texttt{validate\us runner.py}; item~1 of \S\ref{app:audit-notes}), which evaluates the gradient as \texttt{einsum('k,kid->id', einsum('kij,ij->k', A, U @ U.T) - b, matmul(A, U)) * (4/m)}. Figure~\ref{fig:eps0}(a): the onset estimator is row H1 of Table~\ref{tab:ledger-h}, and the two open markers are P44 ($(8,4)$, $\varepsilon_0 = 10^{-2}$) and P42 ($(16,8)$, $\varepsilon_0 = 10^{-6}$). Figure~\ref{fig:eps0}(b): P34 ($\varepsilon_0 = 10^{-10}$) and P30 ($\varepsilon_0 = 10^{-12}$). Column~7 of Table~\ref{tab:scales}: P30/P38. Column~8 of Table~\ref{tab:scales} and the matched-offset $\kappa$ probes in \S\ref{sec:scale-transfer} and \S\ref{sec:discussion}: P43, with decision outcome \texttt{\numKapTrackVerdict}. The initial two-configuration $\kappa$ comparison in \S\ref{sec:scale-transfer}: P37. The low-$\beta_1$ onsets in \S\ref{sec:interventions}, item~2: P45, which runs the P32 ($\beta_1 = 0.5$) and P36 ($\beta_1 = 0$) code unchanged on $s \in \{2.5, \dots, 4.5\}$. Figure~\ref{fig:decay} (left): \texttt{p15\us decay.jsonl}.

\subsection{Protocol-by-protocol audit notes}\label{app:audit-notes}

\textbf{1. Code-path reconciliation on the six-cell scale table (P7 stored vs.\ P38 current-path reruns).}
During audit of the multi-scale runs (\texttt{p7\us scales.jsonl}, stored from an earlier runner prior to a floating-point associative parenthesisation update in the Adam step \texttt{lr * (mhat / (sqrt(vhat) + eps))} vs.\ \texttt{(lr * mhat) / (sqrt(vhat) + eps)}), all six $(V,D)$ cells were re-verified against the current released runner (\texttt{validate\us runner.py}). On the $\numFigAboveCells$ shared above-threshold grid points in Figure~\ref{fig:eps0}(b), the two evaluation paths agree within $\numFigAgreePct\%$ on $\relDM$. On the current-path $D \ge 4$ reruns, the fitted floor is $\varepsilon_{\mathrm{dyn}} = \numEpsDynRerun\times 10^{-9}$ (ratio $\numEpsDynRerunOverStored$ to the stored-path $\numEpsDyn\times 10^{-9}$, RMS $\numRmsDgeRerun$), and at $(8,4)$, $s=6.0$ the current-path $n=5$ geometric mean of $\sqrt{\hat{v}_V}/\alpha$ is $\numKappaCurrentSix\times 10^{-6}$ (ratio $\numKappaStoredOverCurrent\times$ to the stored 3-seed $\numKappa\times 10^{-6}$). Across the six-cell $n=5$ multi-scale table (Table~\ref{tab:scales}), the initial $\numRerunCells$ rerun cells reproduced their earlier audit values to 3 decimal places ($\numRerunReproduced/\numRerunCells$), and the rerun-minus-stored onset differences across all six configurations are:
\begin{itemize}\setlength{\itemsep}{1pt}\setlength{\parskip}{0pt}
    \item $(8,4)$: stored $\numOnsetVEightDFour$, rerun $\numRerunVEightDFour$, shift $\numRerunShiftVEightDFour$ (log-interpolation shift $\numRerunShiftVEightDFourLog$; published $60\text{k}$ LOO half-width $\numLooHalfWidthVEightDFour$).
    \item $(16,4)$: stored $\numOnsetVOneSixDFour$, rerun $\numRerunVOneSixDFour$, shift $\numRerunShiftVOneSixDFour$ (log-interpolation shift $\numRerunShiftVOneSixDFourLog$).
    \item $(16,8)$: stored $\numOnsetVOneSixDEight$, rerun $\numRerunVOneSixDEight$, shift $\numRerunShiftVOneSixDEight$.
    \item $(32,4)$: stored $\numOnsetVThreeTwoDFour$ (log-interpolated $\numOnsetVThreeTwoDFourLog$, LOO width $\numOnsetVThreeTwoDFourLooWidth$, bracketing ratio $\numInterpBracketRatioVThreeTwoDFour\times$), rerun $\numRerunVThreeTwoDFour$ (log-interpolated $\numRerunVThreeTwoDFourLog$), shift $\numRerunShiftVThreeTwoDFour$ (log shift $\numRerunShiftVThreeTwoDFourLog$).
    \item $(32,8)$: stored $\numOnsetVThreeTwoDEight$, rerun $\numRerunVThreeTwoDEight$, shift $\numRerunShiftVThreeTwoDEight$ (the largest shift across the six cells, $|\Delta| = \numRerunMaxShift$, placing $\numRerunOutsideLoo$ of $6$ cells outside the stored 5-seed LOO band).
    \item $(16,2)$: stored $\numOnsetVOneSixDTwo$, rerun $\numRerunVOneSixDTwo$, shift $\numRerunShiftVOneSixDTwo$.
\end{itemize}
Across the six $n=5$ cells, the log-minus-linear interpolation shift has mean $\numInterpShiftMean$ (range $[\numInterpShiftLo, \numInterpShiftHi]$, exceeding the LOO width on $\numInterpCellsExceedLoo$ cell; across the four $600\text{k}$ learning-rate arms, the maximum log-minus-linear shift is $\numLrArmShiftMax$). In addition, all $\numFRerunFiles$ stored-path trajectory files underlying the numbers in this paper were re-run as a whole on the current code path against equivalence bands frozen prior to the run: across the $\numFRerunReads$ registered reads, none changed its verdict ($\numFRerunExact$ exact matches on byte-identical 1-D recurrence files, $\numFRerunOnset$ $60\text{k}$-step onsets within the pre-registered sampling band, the $(16,8)$ saturation-move headline reproduced within band, $\numFRerunSameBranch$ reads on the same verdict branch, and the remaining aligned-decay reads already on the current code path, while $\numFRerunDiffer$ macros in \texttt{numbers.tex} differed in printed value between the stored and swapped chains without changing any verdict).

\textbf{2. Leave-one-seed-out vs.\ 20-seed bootstrap sampling variability (P27--P29).}
At $(8,4)$, $\varepsilon_0 = 10^{-12}$, comparing the published 5-seed LOO half-width ($\numLooHalfWidthVEightDFour$) against the 20-seed ensemble (seed decade 0--9 in P28 giving onsets $\numDecadeOneSixtyK, \numDecadeOneTwoHundredK, \numDecadeOneSixHundredK$ at $60\text{k}, 200\text{k}, 600\text{k}$, vs.\ seed decade 10--19 in P29 giving $\numDecadeTwoSixtyK, \numDecadeTwoTwoHundredK, \numDecadeTwoSixHundredK$, inter-decade gaps $\numDecadeGapSixtyK, \numDecadeGapTwoHundredK, \numDecadeGapSixHundredK$) shows that the 5-of-20 subset SD at $60\text{k}$ ($\numBootSdSixtyK$) is $\numBootOverLoo\times$ the 5-seed LOO half-width, and the 95\% subset range ($\numBootRangeSixtyK$) is $\numBootRangeOverLoo\times$ the LOO band width. Partitioning the 20 seeds into four disjoint 5-seed subsets gives onsets $\numBootDisjointSixtyK$ (spread $\numBootDisjointSpreadSixtyK$) at $60\text{k}$, $\numBootDisjointTwoHundredK$ (spread $\numBootDisjointSpreadTwoHundredK$) at $200\text{k}$, and $\numBootDisjointSixHundredK$ (spread $\numBootDisjointSpreadSixHundredK$) at $600\text{k}$.

\textbf{3. Step-size decay and constant-$\alpha$ floor scaling (P15, P15b, P16, P20, P21).}
In the initial 3-seed $500\text{k}$/1M step-size decay comparison, the baseline final $\relDM$ was $\numBaseFinalFive\times 10^{-2}$ at $s=5.5$ and $\numBaseFinalSix\times 10^{-2}$ at $s=6.0$, versus decayed final $\relDM$ of $\numDecayFinalFive\times 10^{-3}$ ($\numDecayGainFive\times$ gain) and $\numDecayFinalSix\times 10^{-3}$ ($\numDecayGainSix\times$ gain; settled floor ratio $\numFloorRatioFiveHundred\times$ for the $500\text{k}$ drop, pooled null-over-decay ratios $\numNullGainFive\times$ with arms $\numNullGainFiveOneE\times/\numNullGainFiveThreeE\times$ at $s=5.5$, and $\numNullGainSix\times$ with arms $\numNullGainSixOneE\times/\numNullGainSixThreeE\times$ at $s=6.0$). Across the four constant-$\alpha$ arms spanning $\numFloorDecades$ decades ($\alpha \in \{10^{-2}, 3\times 10^{-3}, 10^{-3}, 3\times 10^{-4}\}$), the settled floor scales as $\alpha^{\numAlignFloorExp}$ (LOO range $[\numAlignFloorLooLo, \numAlignFloorLooHi]$, individual folds $\numAlignFloorLooFolds$, local log-log slopes $\numAlignFloorLocalSlopes$ with last-step slope $\numAlignFloorLocalSlopeLast$; compared to $\numFloorExp$ when mixing constant and decayed arms, or $\numFloorExpDecayed$ across decayed arms only). In the $n=5$ aligned-budget $T = 2\text{M}$-step comparison against the constant-$\alpha_1$ null control (\S\ref{sec:interventions}, item 6), the full summary statistics across $s \in \{5.5, 6.0\}$ are:
\begin{center}\scriptsize
\setlength{\tabcolsep}{2pt}
\begin{tabular}{lccccccc}
\toprule
Cell & Med.\ ratio ($\Delta\text{dex}$) & LOO range & LOO ($\ge 2 / \le 0.5$) & Paired (dir.) & Paired range & Mean & Spread (dec./null) \\
\midrule
$s = 6.0$ & $\numAlignRatioSix\times$ ($\numAlignDexSix$) & $[\numAlignLooSixLo, \numAlignLooSixHi]$ & $\numAlignLooSixSchedule / \numAlignLooSixHarmful$ & $\numAlignPairedSix\times$ ($\numAlignPairedSixDir/5$) & $\numAlignPairedSixLo\text{--}\numAlignPairedSixHi\times$ & $\numAlignMeanSix\times$ & $\numAlignDecaySpreadSix\times / \numAlignNullSpreadSix\times$ \\
$s = 5.5$ & $\numAlignRatioFive\times$ ($1/\numAlignWorseFive\times$, $+\numAlignDexFive$) & $[\numAlignLooFiveLo, \numAlignLooFiveHi]$ & $\numAlignLooFiveSchedule / \numAlignLooFiveHarmful$ & $\numAlignPairedFive\times$ ($\numAlignPairedFiveDir/5$) & $\numAlignPairedFiveLo\text{--}\numAlignPairedFiveHi\times$ & $\numAlignMeanFive\times$ & $\numAlignDecaySpreadFive\times / \numAlignNullSpreadFive\times$ \\
\bottomrule
\end{tabular}
\end{center}

\subsection{Dispersion intervals and alternative fits moved from the main text}

Table~\ref{tab:dispersion} collects the leave-one-seed-out intervals, per-seed ranges, and alternative fits that support statements in the main text, which quotes only the headline values.

\begin{table}[!htb]
\centering
\caption{Dispersion intervals and alternative fits behind main-text statements.}
\label{tab:dispersion}
\small
\begin{tabular}{p{0.27\linewidth}p{0.65\linewidth}}
\toprule
Main-text statement & Values \\
\midrule
\S\ref{sec:elimination}, budget test, $n=5$ comparison at $60\text{k}$ & $\Delta\sstar(10^{-12} - 10^{-10}) = \numPathDeltaSixtyK$, LOO $[\numPathDeltaSixtyKLooLo, \numPathDeltaSixtyKLooHi]$; onsets $\numPathOnsetTen$ vs.\ $\numPathOnsetTwelve$ \\
\S\ref{sec:elimination}, $600\text{k}$ onset under log-interpolation & $\numPNineOnsetLog$, with $\numPNineLooAboveLog$ LOO folds above $6.0$ \\
\S\ref{sec:cross-mode}, walls from an unscaled $\kappa\alpha$ & $\numWallKappaAlphaBorrowed$ at $(8,4)$ under the companion lag $\delta_{\mathrm{Adam}} = 0.482$; under the lag calibrated on the $\varepsilon_0$-dominated rungs, $\delta_{\mathrm{branch}} = \numDeltaBranch = 0.482 \numEpsBranchOffset$ (where $\varepsilon_{\mathrm{eff}}/\varepsilon_0 \approx \numEpsBranchFactor$), $\numWallKappaAlphaBranchVEightDFour$ at $(8,4)$ and $\numWallKappaAlphaBranchVOneSixDEight$ at $(16,8)$, with residuals $\numWallKappaAlphaBranchResidVEightDFour$ and $\numWallKappaAlphaBranchResidVOneSixDEight$ \\
\S\ref{sec:scale-transfer}, $(\delta, \varepsilon_{\mathrm{dyn}})$ ridge & $\varepsilon_{\mathrm{dyn}}$ from $\numEpsDynLo\times 10^{-9}$ to $\numEpsDynHi\times 10^{-9}$ over that $\delta$ range; free two-parameter fit $\delta = \numDeltaFree$, $\varepsilon_{\mathrm{dyn}} = \numEpsDynFree \times 10^{-9}$, RMS $\numRmsDgeFree$; at $\delta = 1.0$, $\varepsilon_{\mathrm{dyn}} = \numEpsDynDeltaOne \times 10^{-9}$, RMS $\numRmsDgeDeltaOne$; at $\delta = \numDeltaBranch$, RMS $\numRmsDgeBranch$ \\
\S\ref{sec:scale-transfer}, 5-of-20 subsets at $(8,4)$ & SD $\numBootSdSixHundredK$ at $600\text{k}$; the 95\% subset range at $60\text{k}$ is $\numBootRangeOverLoo\times$ the LOO band width \\
\S\ref{sec:interventions}, item 2, $\beta_1$ slope & without the $\beta_1 = 0.5$ arm: $\numBetaOneSlopeDropOne$ (that arm carries $\numBetaOneLeverage\%$ of the leverage and has per-arm log-ratio spread $\numBetaOnePairSpreadMax$ dex vs.\ median $\numBetaOnePairSpreadTyp$ dex) \\
\S\ref{sec:interventions}, item 2, span of the duty-cycle wall over $\beta_2$ & LOO span $[\numDutySpanLooLo, \numDutySpanLooHi]$, with $\numDutySpanLooCross/\numDutySpanLooN$ folds crossing $0.25$ \\
\S\ref{sec:interventions}, item 3, $\rho$ at $\alpha = 10^{-3}$ & LOO median range $[\numDefaultLrLooLo, \numDefaultLrLooHi]$; per-seed range $\numDefaultLrRhoLo\text{--}\numDefaultLrRhoHi$, a $\numDefaultLrSpread\times$ spread \\
\S\ref{sec:interventions}, item 4, second matrix family & LOO range of $\Delta\sstar$ $[\numFamTwoDeltaLooLo, \numFamTwoDeltaLooHi]$; per-seed range of the rare-row denominator $\numFamTwoRhoLo\text{--}\numFamTwoRhoHi\times \varepsilon_0$ at $\varepsilon_0 = 10^{-8}$ \\
\S\ref{sec:interventions}, item 5, mini-batch & LOO of $\Delta\sstar$: $[\numMbLooLoTwelve, \numMbLooHiTwelve]$ at $B=12$ and $[\numMbLooLoTwentyFour, \numMbLooHiTwentyFour]$ at $B=24$ \\
\bottomrule
\end{tabular}
\end{table}

\subsection{Artifact provenance for \texttt{numbers.tex}}

The macro file \texttt{numbers.tex} is generated deterministically by \texttt{scripts\slsh make\us numbers.py} from the JSONL artifacts in \texttt{scripts\slsh}:
\begin{itemize}\setlength{\itemsep}{1pt}\setlength{\parskip}{0pt}
    \item $\varepsilon_0$ sweeps and multi-budget tables (Figure~\ref{fig:eps0}, Table~\ref{tab:budget-sweep}): \texttt{scripts\slsh leg4\us eps\us 8\us 4.jsonl}, \texttt{scripts\slsh leg1\us eps8\us 8\us 4.jsonl}, \texttt{scripts\slsh p8\us decensor\us 8\us 4.jsonl}, \texttt{scripts\slsh leg5\us eps\us 16\us 8.jsonl}, \texttt{scripts\slsh p42\us eps6\us 16\us 8.jsonl}, \texttt{scripts\slsh p44\us eps2\us 8\us 4.jsonl}, \texttt{scripts\slsh p1\us decensored\us delta.jsonl}, \texttt{scripts\slsh p28\us p9\us seedext.jsonl}, \texttt{scripts\slsh p29\us p9\us decade2.jsonl}, \texttt{scripts\slsh p34\us eps10\us 60k.jsonl}, and \texttt{scripts\slsh p34\us eps10\us 600k.jsonl}.
    \item Warm-started L-BFGS / Gauss--Newton and ULP roundoff tests (\S\ref{sec:elimination}): \texttt{scripts\slsh p2\us newton.jsonl} and \texttt{scripts\slsh p30\us ulp.jsonl}.
    \item 1-D flip orbit, Floquet multipliers, flare period, and collapse tests (\S\ref{sec:mechanism}, Figure~\ref{fig:theory}): \texttt{scripts\slsh p7\us floquet.jsonl}, \texttt{scripts\slsh p40\us floquet\us beta1.jsonl}, \texttt{scripts\slsh p7\us 1d\us phase.jsonl}, \texttt{scripts\slsh p7\us 1d\us rms.jsonl}, \texttt{scripts\slsh p12\us flare\us period.jsonl}, \texttt{scripts\slsh p14\us alias\us factor.jsonl}, \texttt{scripts\slsh p27\us u\us range.jsonl}, and \texttt{scripts\slsh p41\us au\us conjugacy.jsonl}.
    \item Rare-row denominator $\sqrt{\hat{v}_V}$, multi-scale $(V,D)$ transfer, and spectral crossover (\S\ref{sec:mechanism}--\S\ref{sec:validation}, Table~\ref{tab:scales}, Appendix~\ref{app:companion-wall}): \texttt{scripts\slsh p4\us gradnoise.jsonl}, \texttt{scripts\slsh p13b\us kappa\us alpha.jsonl}, \texttt{scripts\slsh p37\us kappa\us 16\us 8.jsonl}, \texttt{scripts\slsh p43\us kappa\us cells.jsonl}, \texttt{scripts\slsh p7\us scales.jsonl}, \texttt{scripts\slsh p7\us trajectory.jsonl}, \texttt{scripts\slsh p16\us sweep.jsonl}, \texttt{scripts\slsh p17\us spectral.jsonl}, \texttt{scripts\slsh p17b\us spectral\us exact.jsonl}, \texttt{scripts\slsh p30\us rerun\us 8\us 4.jsonl}, \texttt{scripts\slsh p30\us rerun\us 16\us 4.jsonl}, \texttt{scripts\slsh p30\us rerun\us 32\us 4.jsonl}, \texttt{scripts\slsh p38\us rerun\us 16\us 8.jsonl}, \texttt{scripts\slsh p38\us rerun\us 32\us 8.jsonl}, and \texttt{scripts\slsh p38\us rerun\us 16\us 2.jsonl}.
    \item Mechanistic interventions ($\beta_1$, RMSProp, default $\alpha=10^{-3}$, weighted factorization, mini-batch, and step-size decay; \S\ref{sec:interventions}, Figure~\ref{fig:decay}): \texttt{scripts\slsh p14\us duty.jsonl}, \texttt{scripts\slsh p14\us beta2\us duty.jsonl}, \texttt{scripts\slsh p15\us decay.jsonl}, \texttt{scripts\slsh p15\us alpha\us floor.jsonl}, \texttt{scripts\slsh p15b\us decay\us aligned.jsonl}, \texttt{scripts\slsh p32\us beta1.jsonl}, \texttt{scripts\slsh p33\us default\us lr.jsonl}, \texttt{scripts\slsh p35\us family.jsonl}, \texttt{scripts\slsh p36\us rmsprop.jsonl}, \texttt{scripts\slsh p45\us beta1\us low.jsonl}, and \texttt{scripts\slsh p39\us minibatch.jsonl}.
\end{itemize}

\end{document}